\documentclass[preprint,12pt]{elsarticle}%#,authoryear

\usepackage{amssymb}
\usepackage{amsmath,amssymb,amsfonts}
\usepackage{algorithmic}
\usepackage{graphicx}
\usepackage{textcomp}
\usepackage{xcolor}
\usepackage{mathtools}
\usepackage{balance}
\usepackage{caption}
\usepackage{subcaption}
\usepackage{enumitem}

\newcommand{\ve}[1]{{\mbox{\boldmath${#1}$}}}

\setcitestyle{square}

\journal{Neurocomputing}

\begin{document}

\begin{frontmatter}



\title{Deep Time Series Models for Scarce Data}


\author{Qiyao Wang, Ahmed Farahat, Chetan Gupta, Shuai Zheng}

\address{Industrial AI Laboratory, Hitachi America, Ltd. R$\&$D \\
Santa Clara, CA, USA \\
$\{$Qiyao.Wang, Ahmed.Farahat, Chetan.Gupta, Shuai.Zheng$\}$@hal.hitachi.com
}

\begin{abstract}
Time series data have grown at an explosive rate in numerous domains and have 
stimulated a surge of time series modeling research. A comprehensive comparison of different time series models, for a considered data analytics task, provides useful guidance on model selection for data analytics practitioners. Data scarcity is a universal issue that occurs in a vast range of data analytics problems, due to the high costs associated with collecting, generating, and labeling data as well as some data quality issues such as missing data. In this paper, we focus on the temporal classification/regression problem that attempts to build a mathematical mapping from multivariate time series inputs to a discrete class label or a real-valued response variable. For this specific problem, we identify two types of scarce data: scarce data with small samples and scarce data with sparsely and irregularly observed time series covariates. Observing that all existing works are incapable of utilizing the sparse time series inputs for proper modeling building, we propose a model called sparse functional multilayer perceptron (SFMLP) for handling the sparsity in the time series covariates. The effectiveness of the proposed SFMLP under each of the two types of data scarcity, in comparison with the conventional deep sequential learning models (e.g., Recurrent Neural Network, and Long Short-Term Memory), is investigated through mathematical arguments and numerical experiments \footnote{This paper is an extended version of the following papers written by the authors \cite{wang2019multilayer, wang2019remaining}.}. 

%In this paper, we propose a way of generalizing Multilayer Perceptrons (MLP) to sparse functional data, where for a given subject there are multiple observations available over time and these observations are sparsely and irregularly distributed within the time range considered. We justify the validity of our algorithm through theoretical arguments. We apply our proptosed method to solving three data challenges, including classification of synthetics curves, prediction of patient's long term survival, and prediction of turbofan engine's remaining time to critical failures. To show the superiority of our algorithm under sparse functional data scenarios, we compare the performance of our model with two alternative common practices, and our method outperforms the baselines in all the numerical studies. 

\end{abstract}




\end{frontmatter}


\section{Introduction}
%\subsection{Relevant work on time series analysis}
Nowadays, time series data that consist of repeated data measurements over a bounded time range have become ubiquitous in numerous domains, such as meteorology, epidemiology, transportation, agriculture, industry, bioinformatics, and the world wide web. Time series data often have intrinsic temporal structures such as auto-correlation, trend, and seasonality. For example, for industrial equipment, the sensor measurements over the lifespan are correlated. Also, the sensor time series often gradually increase/decrease due to performance degradation. When modeling time series data, it is of paramount importance to leverage the internal temporal information to achieve good data analytical performance. 

In the literature, three types of time series approaches have been widely explored, including (1) the classical time series models (e.g., the Auto-Regressive Integrated Moving Average and the autoregressive exogenous models) \cite{hamilton1994time},  (2) the sequential deep learning models such as the Recurrent Neural Network (RNN), the Long Short-Term Memory (LSTM), and the Gated Recurrent Unit (GRU) in the machine learning community \cite{mikolov2010recurrent, connor1994recurrent, chung2014empirical}, and the (3) emerging functional data analysis in the statistical field \cite{ramsay2006functional}. Fundamentally, the classical time series models exploit the autoregressive (AR) and moving average (MA) techniques to encode the dependency of later observations on the prior data and the regression error at previous timestamps. The deep sequential learning models use hidden states to hold the up-to-present memory and recurrently conduct the same transformation on the internal memory and input data along the timestamps to sequentially process the temporal information. Rather than considering time series as a sequence of scalar-valued observations, functional data analysis (FDA) models treat the observed time series data as discrete realizations given rise by a continuous underlying random function of time (i.e., random curve) \cite{ramsay2006functional} and directly analyze a sample of such finitely evaluated random functions. This functional setting naturally accounts for temporal information. There is substantial literature on modeling and estimation for functional data, including functional principal component analysis \citep{castro1986principal, silverman1996smoothed}, regression with functional responses, functional predictors or both \citep{cardot2003spline, cai2006prediction, muller2005time, yao2005functional}, functional classification and clustering \citep{james2003clustering,muller2005functional}, and functional quantile analysis \citep{ cardot2005quantile,ferraty2005conditional, chen2012conditional}.  Ramsay \cite{ramsay2006functional} offers a comprehensive perspective of FDA methods. 

%Different from these two types of methods, functional data analysis (FDA) models treat the observed time series data as discrete realizations given rise by an underlying random function of time (i.e., random curve), rather than a sequence of scalar-valued observations \cite{ramsay2006functional}. 

%

%FDA models use a sample of discretely evaluated random functions

%to estimate the average temporal evolution and to explain the smooth variations across samples. 

The aforementioned time series analytic modeling techniques handle temporal information from different perspectives. Correspondingly, they impose varying requirements on data and achieve different performances under diverse scenarios. A comprehensive comparison of time series models, for a considered data analytics task, provides useful guidance on model selection for data analytics practitioners. Some papers \cite{adebiyi2014comparison, ho2002comparative} conducted comparative studies of RNN and ARIMA for the time series forecasting task, but the literature still lacks a comprehensive comparison between FDA and these approaches. 

In this paper, we focus on the crucial problem of building supervised learning models in time series analysis. In particular, the goal is to build a mathematical mapping from one or several real-valued time series within a bounded period to a discrete class label or a real-valued response variable. Note that we consider the general case where the response variable is different from the temporal covariates. The classical time series models such as ARIMA and the autoregressive exogenous models are infeasible for the considered temporal classification/regression problem, as they are incapable of building mappings for time series to a heterogeneous response. In the other two categories of time series models, several methods are feasible to solve the considered problem, of which the most important two are the sequential learning models \cite{mikolov2010recurrent, hochreiter1997long, chung2014empirical}, and the Functional Multilayer Perceptron (FMLP), a counterpart of the classical MLP for continuous random functions over a continuum such as time series \cite{rossi2005functional, rossi2005representation, wang2019remaining}. 

%Note that the classical time series models such as ARIMA and the autoregressive exogenous models are infeasible for the considered temporal classification/regression problem, as they are incapable of building mappings for time series to a heterogeneous response. 

%build temporal predictive models
%The focus of this paper is on temporal predictive models.

%In this paper, we focus on the problem of

%, i.e., supervised learning with time series inputs. Depending on the data type of the response, the goal of the temporal predictive model is to predict either a discrete class label or a real-valued response, based on time series covariates. 
%Another crucial problem in time series analysis is to build 

\begin{figure*}[htbp]
	\centering
	\begin{subfigure}[t]{1.55in}
		\centering
        \caption{}\label{e1}
        %\vspace{-0.08in}
	\includegraphics[width=42.7mm, height=37mm]{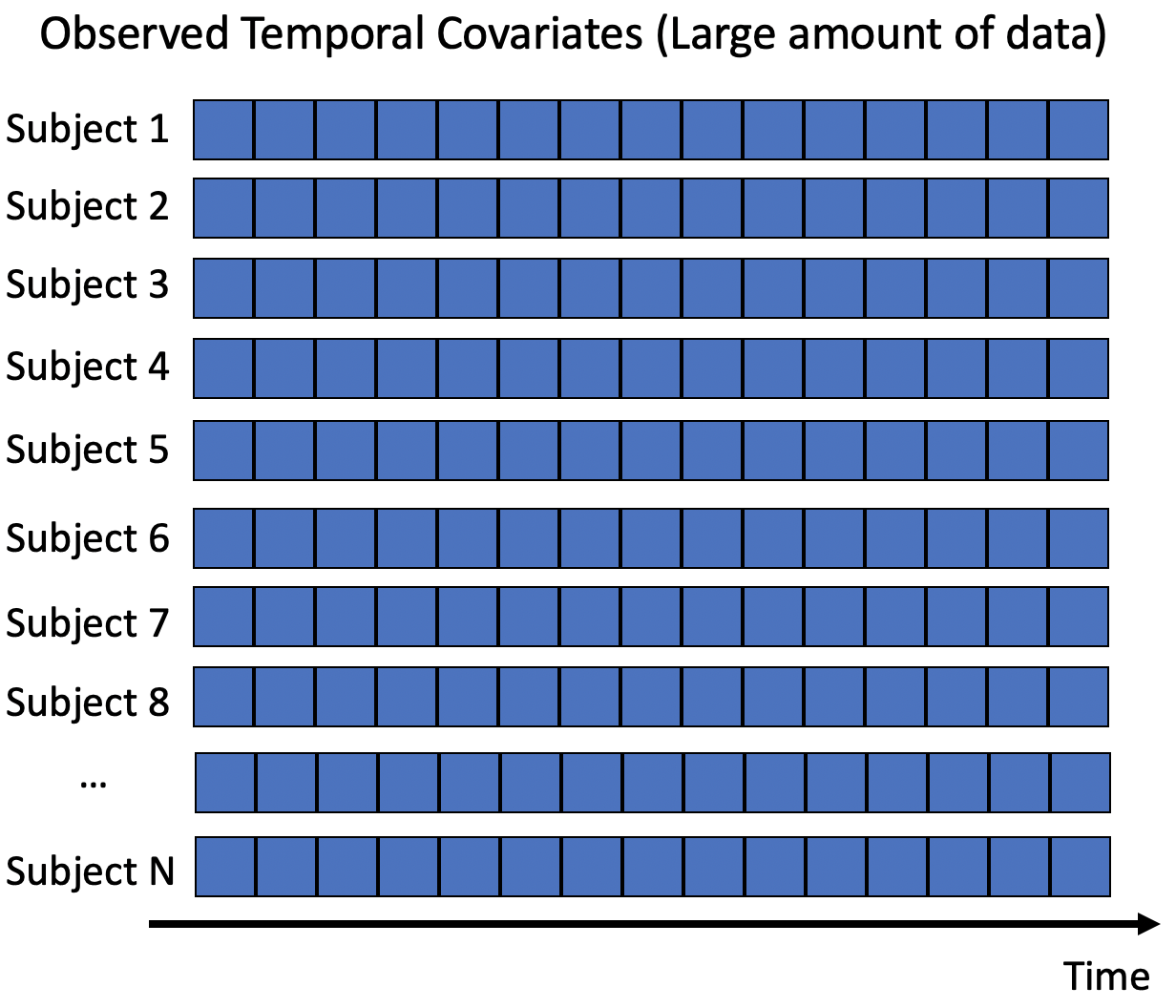}		
	\end{subfigure}
    \quad
	\begin{subfigure}[t]{1.55in}
		\centering
        \caption{}
        \label{e2}
        %\vspace{-0.08in}
		\includegraphics[width=42.7mm, height=37mm]{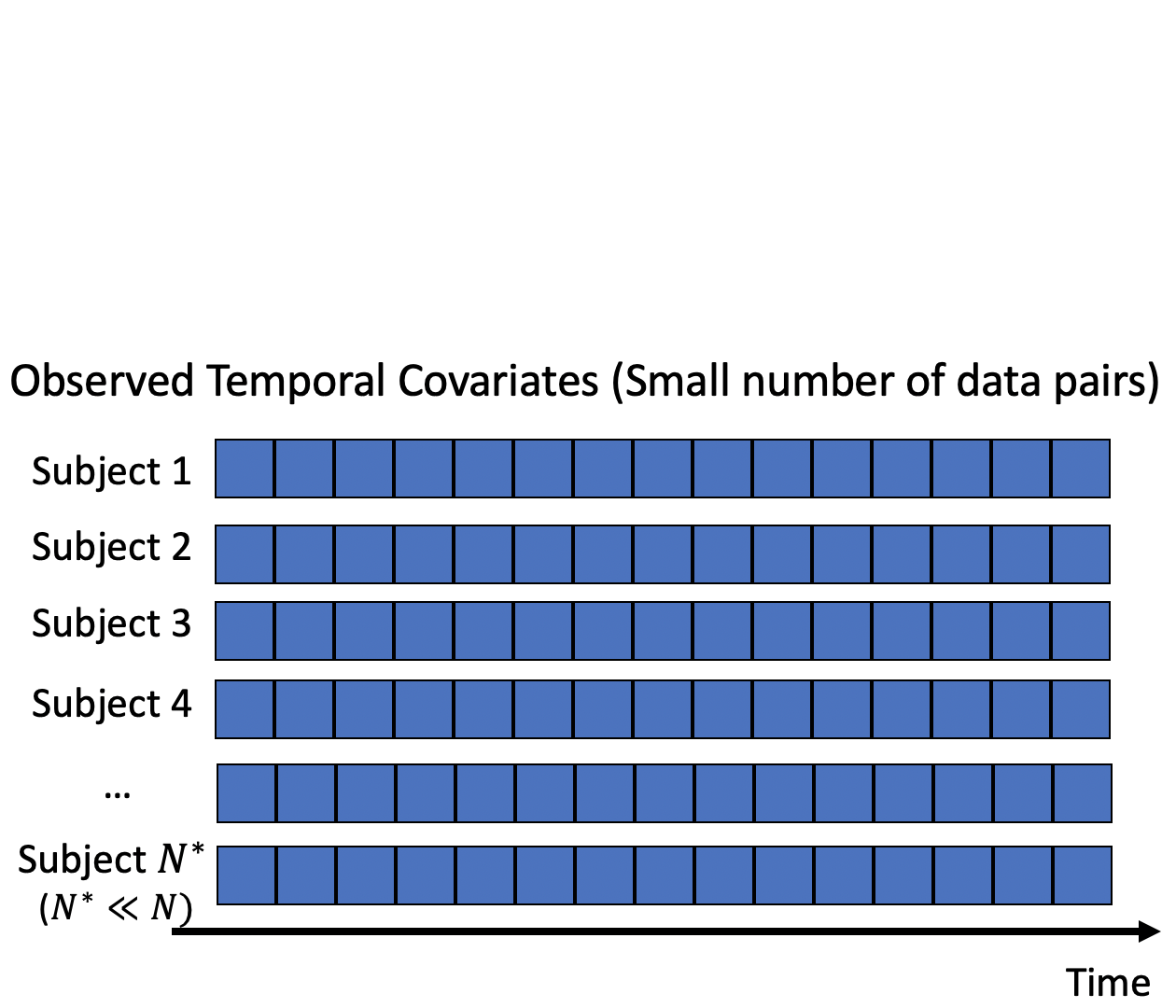} 
	\end{subfigure}
    \quad
	\begin{subfigure}[t]{1.55in}
		\centering
        \caption{}\label{e3}
        %\vspace{-0.08in}
	    \includegraphics[width=42.7mm, height=37mm]{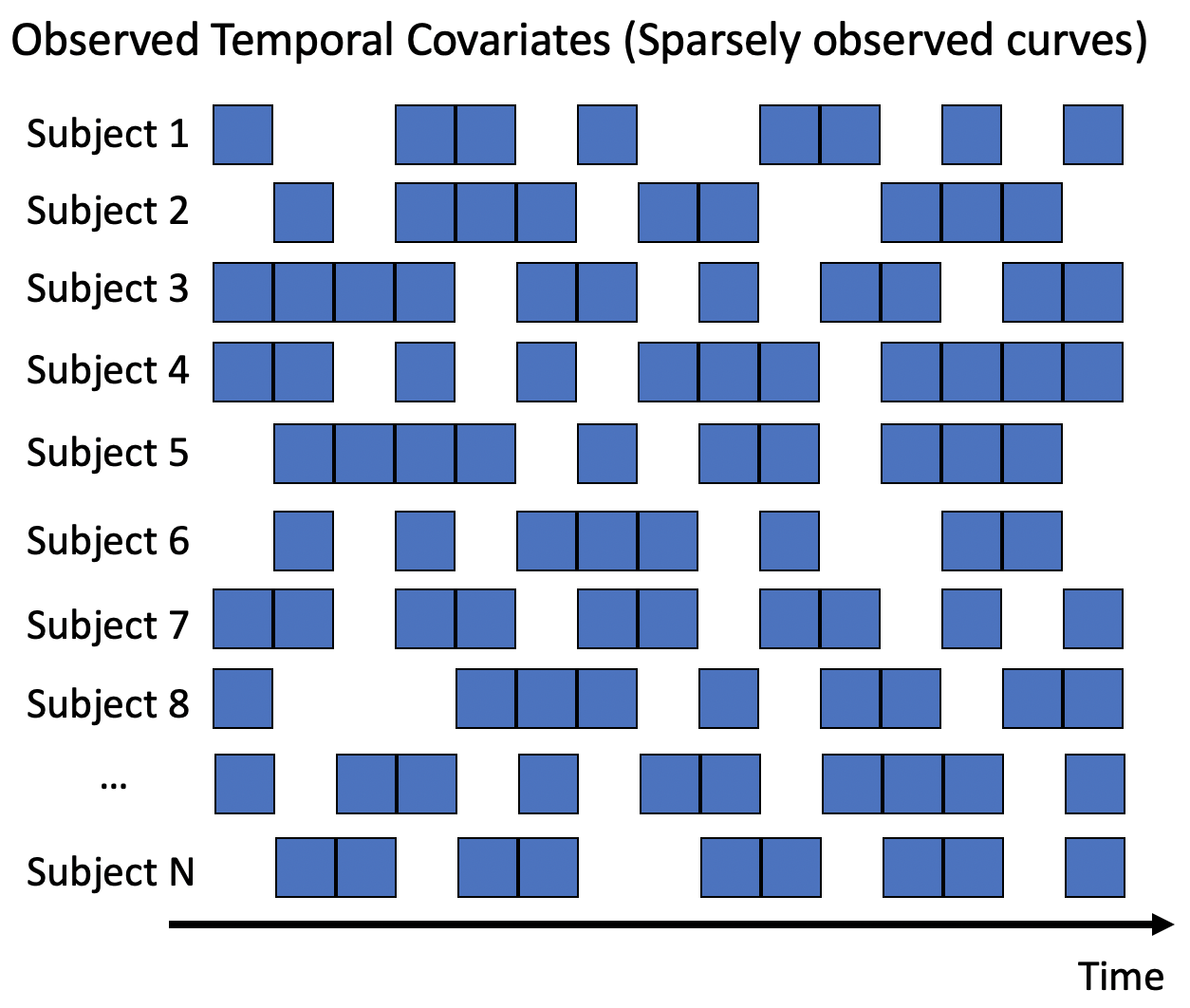}
	\end{subfigure}
	\vspace{-0.05in}
	\caption{Left: An example of large samples. Middle: Scarce data with a limited number of data pairs. Right: Scarce data with sparsely evaluated time series covariates.}\label{lstm12}
	%\vspace{-0.15in}
\end{figure*}

Due to the nature of deep learning models, the sequential and functional deep learning models can both be used to train an end-to-end temporal classification/regression model with reasonable generalizability if we have access to a large number of training samples that cover the diverse variability in data. However, due to the high cost associated with collecting, labeling, storing, processing, and modeling a large amount of training data, building effective deep learning models with a limited amount of samples (i.e., scarce data) is an appealing and meaningful topic in the time series analysis field. In particular to the considered problem, we identify two types of scarce data which can be described as follows. The ideal case is that the number of samples $N$ is sufficiently large and each time series input is densely and regularly observed time series, as shown in Figure \ref{e1}. Scarce data occur when any of these requirements are not satisfied: scarce data with a limited number of samples as illustrated in Figure \ref{e2} and scarce data with sparsely and irregularly evaluated time series covariates as illustrated in Figure \ref{e3}.

This paper first presents a review of the existing sequential and functional temporal predictive models. Observing that the existing method's limitations in dealing with the sparse and irregularly observed covariates, we introduce two sparse functional MLP based on the univariate and multivariate sparse functional principal component analysis \cite{yao2005functional, happ2018multivariate, chiou2014multivariate} \footnote{A preliminary version of this method appears as \cite{wang2019multilayer}.}. The contributions of this paper are summarized as follows:
\begin{enumerate}[leftmargin=*]
    \item We discuss the different types of data scarcity for the temporal classification/regression problem.  
    \item We introduce a new temporal predictive model specially designed to handle scenarios with sparsely and irregularly observed time series inputs.
    %\item We introduce a new temporal predictive models specially designed to handle scenarios with sparsely and irregularly observed time series inputs.
    \item We use mathematical arguments and numerical experiments to investigate each model's feasibility and efficiency in building temporal predictive models under various types of data scarcity.  
\end{enumerate}

%a sparse Functional MLP (SFMLP) for handling scarce data with sparse time series inputs \footnote{A preliminary version of this method appears as \cite{wang2019multilayer}.}. The proposed SFMLP is a generalization of the dense functional MLP \cite{rossi2005functional} equipped with the sparse data Principal Components Analysis through Conditional Expectation (PACE) technique \cite{yao2005functional} to handle scarcity in time series inputs

The rest of the paper is organized as follows. Some preliminaries, including the problem formulation, and review of sequential learning models and the conventional FMLP that only works for dense and regular time series \cite{rossi2005functional, wang2019remaining} are presented in Section \ref{sec2}. The proposed sparse FMLP for sparse time series inputs is described in Section \ref{sec2.4}. The performance of the candidate models under the two types of scarce data scenarios is respectively investigated in Section \ref{sec3} and \ref{sec4}. The paper is concluded in Section \ref{sec5}.

\section{Preliminaries}\label{sec2}

\subsection{Problem formulation}\label{sec2.1}
The goal of temporal predictive models is to build a mapping from multivariate time series covariates to a scalar response variable with good accuracy and generalizability by leveraging the temporal patterns and dependencies. 

Suppose that we have access to $N$ independent training samples. For each subject $i\in \{1,...,N\}$, $R$ features are repeatedly measured within a bounded time range $\mathcal{T}\subseteq \mathbb{R}$. In practice, the measuring timestamps can vary across different covariates and different subjects. As a result, the subject and feature indexes need to be included in the mathematical notations. In particular, the observed data for the $r$-th covariate of subject $i$ is represented by a $M_{i,r}$ dimensional vector $\mathbf{Z}^{(i,r)} = [Z^{(i,r)}_{1}, ...,Z^{(i,r)}_{j},...,Z^{(i,r)}_{M_{i,r}}]^T$, which correspond to observations at timestamps $T^{(i,r)}_{1},...,T^{(i,r)}_{j},...,T^{(i,r)}_{M_{i,r}}$, with $T^{(i,r)}_{j} \in \mathcal{T}$, for $j=1,...,M_{i,r}$. The response variable is $Y_i$, which is binary for temporal classifications and real-valued for temporal regressions. In summary, the observed data are $\{\mathbf{Z}^{(i,1)}, ..., \mathbf{Z}^{(i,R)}, Y_i\}_{i=1}^N$, based on which the temporal predictive models aim at constructing the mapping in Eq.\eqref{setting1}.
\begin{equation}
Y_i = F(\mathbf{Z}^{(i,1)}, ..., \mathbf{Z}^{(i,R)}).\label{setting1}
\end{equation}

The sample size $N$, the number of observations per curve $M_{i,r}$, and the measuring timestamps $T^{(i,r)}_{1},...,T^{(i,r)}_{j},...,T^{(i,r)}_{M_{i,r}}$ jointly determine the level of data availability. A large data scenario depicted in Figure \ref{e1} means that not only $N$ and $M_{i,r}$ are sufficiently large but also the measuring timestamps uniformly cover the time domain $\mathcal{T}$ for all subjects and covariates. Scarce data occur when at least one of these requirements is not satisfied. In particular, scarce data with a limited number of data points happen if the sample size $N$ is small. Whereas, scarce data with sparsely evaluated time series features correspond to situations where $M_{i,r}$ is a small number and/or there exist large gaps among the measuring times $T^{(i,r)}_{1},...,T^{(i,r)}_{j},...,T^{(i,r)}_{M_{i,r}}$, for at least one subject and one covariate. This paper focuses on examining the advantages and disadvantages of several time series models in solving the problem in Eq.\eqref{setting1} given scarce data. We respectively present the candidate models in the remainder of this section.

\subsection{Sequential learning models}\label{sec2.2}
Sequential learning models such as the Recurrent Neural Network (RNN), the Long Short-Term Memory (LSTM), and the Gated Recurrent Unit (GRU) are generalizations of the fully connected MLP that have internal hidden states to process the sequences of inputs \cite{hochreiter1997long}. The key idea is that they employ a series of MLP-based computational cells with the same architecture and parameters to build a directed neural network structure. Any individual cell in the network takes the actual observations at the current index and the hidden states obtained at the previous step (i.e., memory) to produce the updated hidden states that serve as the input for the next computational cell. When the goal is to predict the scalar response associated with the time series, the achieved hidden states at the last index are fed into a nonlinear function to compute the output state. In RNN, each computational cell has one MLP, while there are multiple interacting MLPs in each recurrent unit in LSTM \cite{mikolov2010recurrent, hochreiter1997long}.

The sequential learning models originally designed for text data mining are capable of capturing the order information and the interactions among observations in sequential inputs \cite{mikolov2010recurrent, hochreiter1997long}. Recently, these sequential learning models are also frequently adopted to model time series data \cite{zheng2017long, adebiyi2014comparison, ho2002comparative}. However, when using these models to handle time series data, it is explicitly required that the multivariate inputs are evaluated at an equally-spaced time grid shared by all subjects. This is because the sequential learning models cannot encode the concrete measuring timestamps associated with the individual observations in the time series inputs. Mathematically, in the observed data samples $\{\mathbf{Z}^{(i,1)}, ..., \mathbf{Z}^{(i,R)}, Y_i\}_{i=1}^N$, the covariate vectors $\mathbf{Z}^{(i,r)}$ are of the same length $M$ across all subjects and all covariates. Also, they are evaluated at equally-spaced time grid within time range $\mathcal{T}$, which is denoted as $T_{1},...,T_{j},...,T_{M}$. In practice, data pre-processing procedures such as interpolation are often implemented to obtain the required regular time series when the raw data is sparse and irregular. It is noteworthy that the conventional data pre-processing techniques significantly contaminate the training data when the individual time series are highly sparse and irregular. 

Let's use RNN as an example to present the mathematics behind the sequential learning models. Suppose that $\mathbf{Z}^{(i)}_j=[Z^{(i,1)}_{j},...,Z^{(i,r)}_{j},...,Z^{(i,R)}_{j}]^T$ represents the $R$ dimensional vector containing the $R$ features at time $T_{j}$. Let the number of hidden units in MLP be $L_{\text{RNN}}$ and the $L_{\text{RNN}}$ dimensional hidden state at time $T_{j}$ be $\mathbf{h^{(i)}_j}$. For $j=0,...,M$, the following calculation is recursively conducted
\begin{equation}\label{SeqDL}
\mathbf{h^{(i)}_j}=U_{\text{act1}} (\mathbf{W_{hh}}\mathbf{h^{(i)}_{j-1}} + \mathbf{W_{hz}}\mathbf{Z}^{(i)}_j),
\end{equation}
where $\mathbf{W_{hh}}$ is a $L_{\text{RNN}}$ by $L_{\text{RNN}}$ dimensional parameter matrix, and $\mathbf{W_{hz}}$ is a $L_{\text{RNN}}$ by $R$ dimensional parameter matrix, and $U_{\text{act1}}(\cdot)$ is a non-linear activation function. Let $\mathbf{W_{yh}}$ denote the parameter matrix that associates the last hidden state $\mathbf{h^{(i)}_{M}}$ with the response variable $Y_i$. In the output layer, the output is computed by
\begin{equation}\label{SeqDL2}
Y_i= U_{\text{act2}}(\mathbf{W_{yh}}\mathbf{h^{(i)}_{M}}),
\end{equation}
where $U_{\text{act2}}(\cdot)$ is a non-linear activation function.

\subsection{Functional data analysis and dense functional MLP}\label{sec2.3}
Functional data analysis (FDA) is an emerging branch in statistics that specializes in the analysis and theory of data dynamically evolves over a continuum. In general, FDA deals with data subjects that can be viewed as a functional form $X_i(t)$ over a continuous index $t$. Frequently encountered FDA-type data include time series data, tracing data such as hand-writings, and image data \cite{ramsay2006functional}. 

From the FDA modeling perspective, the $M_{i,r}$ observations of the $r$-th time series feature of subject $i$ (i.e., $\mathbf{Z}^{(i,r)}$) are treated as discretized realizations from a continuous underlying curve $X^{(i,r)}(t)$ contaminated with zero-mean random errors. 
\begin{equation}
Z^{(i,r)}_{j}=X^{(i,r)}(T^{(i,r)}_{j}) + \epsilon_{i,r,j}.\label{setting1.5}
\end{equation}
FDA predictive models directly handle the continuous time series features and solve the problem defined in Eq.\eqref{setting1} by learning 
\begin{equation}
Y_i = F(X^{(i,1)}(t), ...., X^{(i,R)}(t)).\label{setting2}
\end{equation}

Under certain assumptions on the smoothness of the underlying random functions $X^{(i,r)}(t)$ (e.g., continuous second derivatives exist), the conventional functional linear classification or regression models \cite{ramsay2006functional, yao2010functional} assume and learn the unknown real-valued parameters in a linear-formed mapping 
\begin{equation} \label{flm}
    Y_i = b + \sum_{r=1}^{R} \int_{t\in \mathcal{T}}W_{r}(\ve{\beta}_{r},t)X^{(i,r)}(t)dt,
\end{equation}
where $b\in \mathbb{R}$ is the unknown intercept, $\ve{\beta}_{r}$ is a finite-dimensional vector that quantifies the parameter function $W_{r}(\ve{\beta}_{r},t)$, and $\int_{t\in \mathcal{T}}W_{r}(\ve{\beta}_{r},t)X^{(i,r)}(t)dt$ is a generalization of vector inner product to $L^2(t)$ space and it aggregates the time-varying impact of the time series input on the response. Given Eq.\eqref{flm}, it can be seen that, unlike the sequential learning models, the functional models do not require equally-spaced time series observations and can be effectively trained end-to-end as long as the integral can be consistently approximated based on the actual observations $\mathbf{Z}^{(i,r)}$.

\begin{figure}
\centering
\includegraphics[width=95mm]{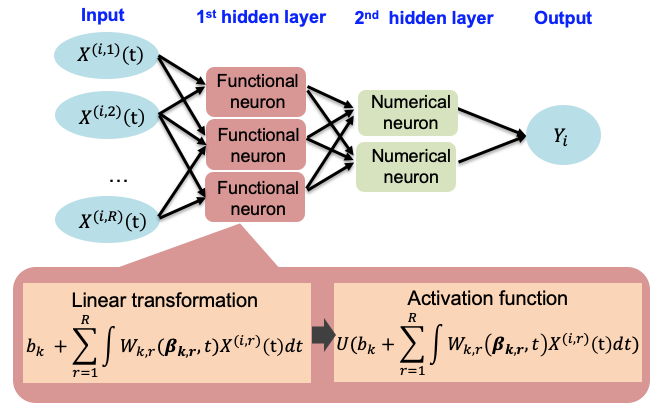}
  \caption{The architecture of a functional MLP with three functional neurons on the first layer and two numerical neurons on the second layer \cite{wang2019remaining}.}
  \label{FNN_exp}
\end{figure}

To capture more complex relationships, a functional MLP that embeds the linear calculation in Eq.\eqref{flm} into the network structure of the conventional MLP \cite{pal1992multilayer} is introduced by \cite{rossi2002functional, rossi2005representation} and later explored further by \cite{wang2019remaining}. In particular, functional MLP proposed a new functional neuron that consists of the linear transformation in Eq.\eqref{flm} and an additional non-linear activation step, as shown in Figure \ref{FNN_exp}. To build functional MLP, multiple functional neurons that take functional inputs and calculates a numerical output are placed on the first layer. The outputs from the functional layer are supplied into subsequent numerical neuron layers whose inputs and outputs are both scalar values, for further manipulations till the output layer that holds the response variable. An example FMLP with three functional neurons on the first layer and two numerical neurons on the second layer is illustrated in Figure \ref{FNN_exp} \cite{wang2019remaining}. 

For the simplicity of mathematical notations, let's consider the case where $K$ functional neurons in the first layer and one numerical neuron in the second layer. Let $U_k(\cdot)$ be an activation function from $\mathbb{R}$ to $\mathbb{R}$, $a_k, b_k \in \mathbb{R}$ for $k=1,..,K$. The weight function for the $r$-th functional feature in the $k$-th functional neuron is assumed to be quantified by a finite dimensional unknown vector $\ve{\beta}_{k,r}$ and is denoted as $W_{k,r}(\ve{\beta}_{k,r}, t)$ for $k=1,...,K$ and $r=1,..,R$. Let $\ve{\beta}=[\ve{\beta}_{1,1},...,\ve{\beta}_{1,R},...,\ve{\beta}_{K,1},....,\ve{\beta}_{K,R}]^T$, $\mathbf{X^{(i)}}=[X^{(i,1)}(t), ...., X^{(i,R)(t)}]^T$. Then the scalar output of the first layer is 
\begin{equation}
H(\mathbf{X^{(i)}}, \ve{\beta})=\sum_{k=1}^{K}a_k U_k(b_k + \sum_{r=1}^{R} \int_{t\in \mathcal{T}}W_{k,r}(\ve{\beta}_{k,r}, t)X^{(i,r)}(t)dt).\label{FN1}
\end{equation}

For more details about the functional MLP including the specifications of the weight functions $W_{k,r}(\ve{\beta}_{k,r}, t)$ in the functional neurons, and assumptions and implementations of the gradient descent based training procedure, we list the literature \cite{rossi2002functional, wang2019remaining} as good references. It is noteworthy that this conventional functional MLP requires that the number of observations per curve is large enough and the individual observations are relatively regularly spaced (i.e., Figure \ref{e2} and \ref{e3}), as stated in Theorem 1 in \cite{rossi2002functional}. In the next section, we propose an effective way of generalizing functional MLP to sparsely and irregularly observed time series inputs.

\section{Proposed sparse functional MLP}\label{sec2.4}
When there is a limited amount of irregularly-spaced data available per feature curve, as shown in Figure \ref{e3}, the existing models described in the previous section are no longer feasible solutions. The deep sequential models need dense and regular observations to model the temporal information through the recurrent network structure. A common practice is to conduct data interpolation to obtain the required dense regular data, however, the conventional interpolation techniques such as the cubic B-spline and the Gaussian process regression often produce biased curve estimates. Analogously, the numerical integration calculation in the conventional FMLP is problematic under sparse data scenarios. In this section, we propose a novel algorithm for generalizing Multilayer Perceptron (MLP) to handle sparse functional data, wherein for a given subject there are multiple observations available over time and these observations are sparsely and irregularly distributed within the considered time range.

\subsection{Sparse functional MLP based on univariate PACE}\label{sec2.4.1}
To derive and define a functional neuron that is calculable for sparse functional data, we propose to go one step further than the existing functional neuron in Eq.\eqref{FN1} with the help of the functional principal component analysis \cite{hall2006properties, silverman1996smoothed, yao2005functional}. Let the $r$-th feature be a random process with an unknown mean function $\mu_r(t)$ and an unknown covariance function $G_r(t,t^\prime)$, $t, t^\prime \in \mathcal{T}$. Mathematically, the non-increasing eigenvalues $\{\lambda_{r,p}\}_{p=1}^\infty$ and the corresponding eigenfunctions $\{\phi_{r,p}(t)\}_{p=1}^\infty$ are solutions of 
\begin{equation}
\lambda \phi(t) = \int_{t^\prime \in \mathcal{T}} G_r(t,t^\prime)\phi(t^\prime)dt^\prime.\label{sparse1}
\end{equation}
The eigenfunctions are orthonormal in the sense that $\int_{t\in \mathcal{T}} \phi_{r,p}(t)\phi_{r,p^\prime}(t)dt=0$ for $p\neq p^\prime$ and $\int_{t\in \mathcal{T}} \phi_{r,p}^2(t)dt=1$. Based on the orthonormality of the eigenfunctions, the $r$-th feature of subject $i$ can be represented as 
\begin{equation}
X^{(i,r)}(t) = \sum_{p=1}^\infty \eta_{i,r,p} \phi_{r,p}(t),\label{sparse2}
\end{equation}
where $\eta_{i,r,p}= \int_{t\in \mathcal{T}}X^{(i,r)}(t)\phi_{r,p}(t) dt$. A common practice for basis expansion-based methods in FDA \cite{pomann2015two,muller2005functional} is to truncate the expansion at the first several directions. This practice is supported by the fact that the core information regarding $X^{(i,r)}(t)$ is mostly captured by the first several basis functions when the curve is smooth in a certain degree. Strict theoretical proofs can be found in \cite{yao2005functional}. For the $r$-th feature, let's truncate at the first $P_r$ dimensions and plug the $X^{(i,r)}(t)$'s approximated representation into the FMLP in Eq.\eqref{FN1}, we have 
\begin{multline}
\label{sparse33}
H(\mathbf{X^{(i)}}, \ve{\beta}) \approx \sum_{k=1}^{K}a_k U_k([b_k + \sum_{r=1}^{R} \int_{t\in \mathcal{T}}W_{k,r}(\ve{\beta}_{k,r}, t) \sum_{p=1}^{P_r} \eta_{i,r,p} \phi_{r,p}(t)dt]).
\end{multline}
Practically, $P_r$ can be selected using the fraction of variance explained approach, AIC or BIC criterion based approach, or the leave-one-curve-out cross-validation method \cite{peng2009geometric}. 

Given Eq.\eqref{sparse33}, it can be seen that the model still cannot directly go through as we cannot consistently estimate $\eta_{i,r,p}=\int_{t\in \mathcal{T}}X^{(i,r)}(t)\phi_{r,p}(t) dt$ from the sparse observations $\mathbf{Z}^{(i,r)}$. Borrowing the idea from sparse data Principal Components Analysis through Conditional Expectation (PACE) \cite{yao2005functional}, we propose to estimate $\eta_{i,r,p}$ by its best linear unbiased predictor, $E[\eta_{i,r,p}|\mathbf{Z}^{(i,r)}]$. This is a reasonable choice to estimate $\eta_{i,r,p}$, because if we take the randomness of measuring time $t$ into account, then 
\begin{equation}
E_t[E[\eta_{i,r,p}|\mathbf{Z}^{(i,r)}]] = E[\eta_{i,r,p}]. \label{sparse4}
\end{equation}
That is to say the random quantities $E[\eta_{i,r,p}|\mathbf{Z}^{(i,r)}]$ and $\eta_{i,r,p}$ share the same expectation. Motivated by a special case where the observations $\mathbf{Z}^{(i,r)}$ and the random errors $\epsilon_{i,r,j}$ are jointly Gaussian distributed, given Eq.\eqref{setting1.5} and \eqref{sparse2}, we can get the explicit formula for $E[\eta_{i,r,p}|\mathbf{Z}^{(i,r)}]$,
\begin{multline}
\label{sparse5}
E[\eta_{i,r,p}|\mathbf{Z}^{(i,r)}]  =\gamma_{r,p} + 
\lambda_{r,p}\ve{\phi}_{i,r,p}^T{[\ve{\Phi}_{i,r}\text{diag}(\ve{\lambda}_r)\ve{\Phi}_{i,r}^T+\sigma_r^2 I]}^{-1}(\mathbf{Z}^{(i,r)}-\ve{\mu}_{i,r}), 
\end{multline}
where $\gamma_{r,p}=\int \phi_{r,p}(t)\mu_{r}(t)\,dt$, $\ve{\lambda}_r=[\lambda_1,...,\lambda_{P_r}]^T$, and $\ve{\phi}_{i,r,p}$ is the eigenfunction $\ve{\phi}_{r,p}(t)$ evaluated at the $M_{i,r}$ observing time points, i.e., $\ve{\phi}_{i,r,p}=[\phi_{r,p}(t_{i,r,1}),...,\phi_{r,p}(t_{i,r,M_{i,r}})]^T$. $\ve{\Phi}_{i,r}$ is a $M_{i,r} \times {P_r}$ matrix, with the $p$-th column being $\ve{\phi}_{i,r,p}$. $\sigma_r$ is the standard deviation of the random noise $\epsilon_{i,r,j}$. By plugging  Eq.\eqref{sparse5} back into Eq.\eqref{sparse33}, we achieve the output of the first layer of our \textbf{\textit{proposed sparse functional MLP}} with $K$ neurons,
\begin{multline}
\label{sparse55}
\tilde{H}(\mathbf{X^{(i)}}, \ve{\beta})=\sum_{k=1}^{K}a_k U_k([b_k + \sum_{r=1}^{R} \int_{t\in \mathcal{T}}W_{k,s}(\ve{\beta}_{k,r}, t)\sum_{p=1}^{P_r} E[\eta_{i,r,p}|\mathbf{Z}^{(i,r)}]\phi_{r,p}(t)dt]).
\end{multline}
The sparse functional neuron in the above equation plays the same role as the functional neuron in Eq.\eqref{FN1} for the conventional functional MLP \cite{rossi2002functional}. 

Before training the proposed sparse functional MLP model, we need to first estimate the unknown values in Eq.\eqref{sparse33} and Eq.\eqref{sparse5}, including the eigenfunctions $\phi_{r,p}(t)$, eigenvalues $\lambda_{r,p}$, standard deviation of random error $\sigma_r$. They can be estimated using the restricted maximum likelihood estimation in \cite{peng2009geometric}, the local linear smoothing based method in \cite{yao2005functional}, or the EM algorithm in \cite{james2000principal}. Since This is not the main focus in this paper, we skip the details. The point is that our algorithm needs consistent estimations for the functional components. All the three methods have been proved to produce consistent estimates for these eigen components for sparse functional data and they can be used to perform the estimation step. In our numerical experiments in Section \ref{sec4}, we used the local linear smoothing based method in \cite{yao2005functional}. Let's denote the estimated values as $\hat{\phi}_{r,p}(t)$, $\hat{\lambda}_{r,p}$ and $\hat{\sigma}_r$. Then we have the empirical counterparts for Eq.\eqref{sparse33} and Eq.\eqref{sparse5}, denoted as $\hat{\tilde{H}}(\mathbf{X^{(i)}}, \ve{\beta})$ and $\hat{E}[\eta_{i,r,p}|\mathbf{Z}^{(i,r)}]$.

After the estimation step, similar to the dense functional MLP in \cite{rossi2002functional}, we propose to use gradient based algorithms to train our sparse functional MLP model. The forward propagation step can go through as follows. First, in the functional neurons, the integral $\int_{t\in \mathcal{T}}W_{k,r}(\ve{\beta}_{k,r}, t)\sum_{p=1}^{P_r}\hat{E}[\eta_{i,r,p}|\mathbf{Z}^{(i,r)}] \hat{\phi}_{r,p}(t)dt$ is approximated by the numerical integration techniques. First layer's output $\hat{\tilde{H}}(\mathbf{X^{(i)}}, \ve{\beta})$ can then calculated by the formula in Eq.\eqref{sparse33}. The forward propagation calculation in subsequent numerical layers is straightforward. In the backward propagation step, the partial derivatives from the output layer up to the second hidden layer (i.e., the numerical layer after the functional neuron layer) can be easily calculated as before. Whereas, it is essential to ensure that the partial derivatives of the values at the second layer (i.e., $\hat{\tilde{H}}(\mathbf{X^{(i)}}, \ve{\beta})$) with respect to the parameters $\ve{\beta}$ exist. This requires that $\partial W_{k,r}(\ve{\beta}_{k,r},t)/\partial \beta_{k,r,q}$ exists almost everywhere for $t\in \mathcal{T}$. Under this assumption, $\partial \hat{\tilde{H}}(\mathbf{X^{(i)}}, \ve{\beta})/\partial \beta_{k,r,q}$ for any $k=1,...,K;r=1,..,R; q=1,...,Q_r$ can be estimated using numerical approximations of the following quantity
\begin{multline}
\label{sparse6}
\frac{\partial\hat{\tilde{H}}(\mathbf{X^{(i)}}, \ve{\beta})}{\partial \beta_{k,r,q}} \approx
a_k U_k^\prime([b_k + \sum_{r=1}^{R} \int_{t\in \mathcal{T}}W_{k,r}(\ve{\beta}_{k,r}, t)\sum_{p=1}^{P_r} \hat{E}[\eta_{i,r,p}|\mathbf{Z}^{(i,r)}]\hat{\phi}_{r,p}(t)dt])\\ 
\times \int_{t\in \mathcal{T}}\frac{\partial W_{k,r}(\ve{\beta}_{k,r}, t)}{\partial \beta_{k,r,q}}\sum_{p=1}^{P_r} \hat{E}[\eta_{i,r,p}|\mathbf{Z}^{(i,r)}]\hat{\phi}_{r,p}(t)dt.
\end{multline}

To justify the validity of our proposal, we have provided brief arguments regarding the consistency of using estimated values as well as the equivalence between our sparse and the proposed dense MLP under dense regular data scenarios. 

\subsection{Sparse functional MLP based on multivariate FPCA}\label{sec2.4.2}
The functional neurons in Eq.\eqref{sparse33} (i.e., an equivalent of the dense functional neuron in Section \ref{sec2.3}) and Eq.\eqref{sparse55} (i.e., the proposed sparse functional neuron) are based on the univariate functional principal component analysis. When separately conducting FPCA, the joint variations among the $R$ variables $\{X^{(i,1)}(t), ...., X^{(i,R)(t)}\}$ are not captured, which makes the random scores from different variables (i.e, $\eta_{i,r,p}$ and $E[\eta_{i,r,p}|\mathbf{Z}^{(i,r)}]$) being correlated and causes multicollinearity issues during modeling. An example of correlated random scores are illustrated in Figure \ref{corr_socre} in Section \ref{sec4}. To overcome this issue, we propose functional neural networks based on the multivariate FPCA \citep{chiou2014multivariate, happ2018multivariate} that are described as follows. 

Let $\mathbf{W}_{k,r}(\ve{\beta}_{k}, t) = [W_{k,1}(\ve{\beta}_{k,r}, t),...,W_{k,r}(\ve{\beta}_{k,R}, t)]^T$, then Eq.\eqref{FN1} can be written as in the following vector format
\begin{equation}
\label{FN1_m}
H(\mathbf{X^{(i)}}, \ve{\beta}) = \sum_{k=1}^{K}a_k U_k([b_k + \int_{t\in \mathcal{T}}\mathbf{W}_{k,r}(\ve{\beta}_{k}, t)^T \mathbf{X}^{(i)}(t)]).
\end{equation}
Let's denote the $R\times R$ matrix that quantifies the covariance of each variable and the joint variation between variables as $\mathbf{G}(t, t^\prime)$, with the $(r,r^\prime)$-th element being $G_{r,r^\prime}(t, t^\prime)=\text{Cov}(X^{(i,r)}(t), X^{(i,r^\prime)}(t^\prime))$. According to the multivariate FPCA, there exists a set of $R$ dimensional orthonormal eigenfunction vectors $\tilde{\ve{\phi}}_p(t)=[\tilde{\phi}_{1,p}(t),...,\tilde{\phi}_{R,p}(t)]^T$, for $p=1,...,\infty$, such that 
\begin{equation} \label{mfpca1}
\int \mathbf{G}(t, t^\prime) \tilde{\ve{\phi}}_p(t^\prime) d t^\prime = \tilde{\lambda}_p \tilde{\ve{\phi}}_p(t),  \text{with } \lim_{p\rightarrow \infty}\tilde{\lambda}_p = 0, 
\end{equation}
where $\tilde{\lambda}_p \in \mathbb{R}$ is the eigenvalue corresponding to the $p$-th eigenfunction vector $\tilde{\ve{\phi}}_p(t)$. Accordingly, it has been shown that the $R$ dimensional data $\mathbf{X}^{(i)}(t)$ can be represented by 
\begin{equation} \label{mfpca2}
\mathbf{X}^{(i)}(t) = \sum_{p=1}^{\infty} \tilde{\eta}_{i,p}\tilde{\ve{\phi}}_p(t) \approx \sum_{p=1}^{P} \tilde{\eta}_{i,p}\tilde{\ve{\phi}}_p(t), 
\end{equation}
where $\tilde{\eta}_{i,p}=\sum_{r=1}^R \int X^{(i,r)}(t)\tilde{\phi}_{r,p}(t)dt$. Comparing the separate FPCA in Section \ref{sec2.4.1} with the multivariate FPCA in Eq.\eqref{mfpca1}~\eqref{mfpca2}, it can be seen that univariate FPCA is a special case of the multivariate FPCA that assumes zero joint variation between variables. Theoretically, we expect the functional regression models in this section to outperform those in Section \ref{sec2.4.1}. The magnitude of improvement is affected by the size of joint variation between variables and the complexity of the underlying mapping. 

Given Eq. \eqref{FN1_m}\eqref{mfpca1} \eqref{mfpca2}, when all the variables $\{X^{(i,r)}(t), i=1,...,N; r=1,...,R\}$ are densely and regularly evaluated, the multivariate functional neuron is defined as follows
\begin{multline}
\label{sparse33_m}
H_{M}(\mathbf{X^{(i)}}, \ve{\beta}) \approx \sum_{k=1}^{K}a_k U_k([b_k +  \int_{t\in \mathcal{T}} \mathbf{W}_{k,r}(\ve{\beta}_{k}, t)^T \sum_{p=1}^{P} \tilde{\eta}_{i,p} \tilde{\ve{\phi}}_p(t)dt]).
\end{multline}
For sparsely evaluated data, analogous to Eq.\eqref{sparse33}, the sparse multivariate function neuron is 
\begin{equation*}
\resizebox{1.0\hsize}{!}{
\label{sparse33_ms}
$H_{M}(\mathbf{X^{(i)}}, \ve{\beta}) \approx \sum_{k=1}^{K}a_k U_k([b_k +  \int \mathbf{W}_{k,r}(\ve{\beta}_{k}, t)^T \sum_{p=1}^{P} E[\tilde{\eta}_{i,p}|\mathbf{Z}^{(i,1)},...,\mathbf{Z}^{(i,R)}] \tilde{\ve{\phi}}_p(t)dt])$
}
\end{equation*}

\begin{equation}
\label{sparse33_ms2}
E[\tilde{\eta}_{i,p}|\mathbf{Z}^{(i,1)},...,\mathbf{Z}^{(i,R)}]=\sum_{r=1}^R E[\tilde{\eta}_{i,p}|\mathbf{Z}^{(i,r)}].
\end{equation}
Note that $E[\tilde{\eta}_{i,p}|\mathbf{Z}^{(i,r)}]$ is calculated by replacing $\phi_{r,p}(t)$ and $\lambda_{r,p}$ with $\tilde{\phi}_{r,p}(t)$ and $\tilde{\lambda}_{p}$ respectively. Based on the multivariate function neurons discussed above, we propose multivariate FMLP by embedding these neurons in the architecture in Figure \ref{FNN_exp}. 

Next, we describe how to numerically implement the multivariate FMLP as follows. The core theoretical result in \cite{happ2018multivariate} is that there is an analytical relationship between the univariate FPCA and the multivariate FPCA, which implied that $\tilde{\phi}_{r,p}(t)$ and $\tilde{\lambda}_{p}$ can be calculated by estimating $\phi_{r,p}(t)$ and $\lambda_{r,p}$ for all $r=1,...,R$. In particular, let the estimated score from univariate FPCA be $\hat{s}_{i,r,p}=\hat{\eta}_{i,r,p}$ or $\hat{s}_{i,r,p}=\hat{E}[\eta_{i,r,p}|\mathbf{Z}^{(i,r)}]$, for $i=1,...,N; r=1,...R; p=1,...,P_r$. Let $P_+=\sum_{r=1}^R P_r$ and $\ve{\Xi}$ is a $P_+ \times P_+$ consisting of blocks $\ve{\Xi^{(rr^\prime)}} \in \mathbb{R} ^{P_{r}\times P_{r^\prime}}$ with the $(p, p^\prime)$-th entry being
\begin{equation}
\begin{split}
\Xi_{pp^\prime}^{(rr^\prime)} & = \text{Cov}(\hat{s}_{i,r,p}, \hat{s}_{i,r^\prime,p^\prime})\\
& = \frac{1}{N-1}\sum_{i=1}^N (\hat{s}_{i,r,p} - \bar{\hat{s}}_{i,r,p} )(\hat{s}_{i,r^\prime,p^\prime} - \bar{\hat{s}}_{i,r^\prime,p^\prime} )
\end{split}
\end{equation}
Let's conduct eigen decomposition on matrix $\ve{\Xi}$ and denote the $p$-th eigenvector as $\mathbf{c}_p$. Note that $\mathbf{c}_p$ can be considered as a vector consisting of $R$ blocks, with the $r$-th block being denoted as $[\mathbf{c}_p]^{(r)} \in \mathbb{R}^{P_r}$. According to the proposition in \cite{happ2018multivariate}, we can estimate $\tilde{\phi}_{r,p}(t)$ by
\begin{equation}
\hat{\tilde{\phi}}_{r,p}(t) = \sum_{m=1}^{P_r}[\mathbf{c}_p]_m^{(r)} \hat{\phi}_{r,m}(t),
\end{equation}
where $\hat{\phi}_{r,m}(t)$ is the achieved eigenfunction from conducting univariate FPCA on the $r$-th variable. The joint eigenvalue  $\tilde{\lambda}_{p}$ is the same as the eigenvalue of matrix $\ve{\Xi}$. 

In summary, we propose two types of extensions of the FMLP in \cite{rossi2005functional, wang2019remaining}. The model in Section \ref{sec2.4.1} equipped with the sparse functional principal component analysis is proposed to handle sparse data cases. The models in Section \ref{sec2.4.2} are generalizations of the FMLP to explicitly account for the correlations among variables. The FMLPs using the conventional FPCA are more straightforward to implement, while the multivariate FMLPs are expected to be more accurate, especially when the correlations among the variables are large. 

%As mentioned above, the multivariate FMLP models in Section \ref{sec2.4.2}  

%Random quantity $E[\eta_{i,r,p}|\mathbf{Z^{(i,r)}}]$ in Eq.\eqref{sparse55} is not the same as $\eta_{i,r,p}$ in Eq.\eqref{sparse33}. However, it is the best linear unbiased predictor for $\eta_{i,r,p}$, which provides an accurate enough solution when $\eta_{i,r,p}=\int_{t\in \mathcal{T}}X^{(i,r)}(t)\phi_{r,p}(t) dt$ is not computable from the raw observations $\mathbf{Z^{(i,r)}}$. 

\section{Scarce data with limited number of samples}\label{sec3}
In this section, we present a comparative study of the sequential learning models and the regular FMLP (equivalently, the proposed sparse FMLP) under scenarios where the sample size is small, i.e., Figure \ref{e2}. We first theoretically compare the minimum sample size required by each model by calculating the number of parameters. We also discuss their feasibility and efficiency in dealing with two different types of time series inputs. Finally, we conduct numerical experiments to demonstrate their performance in solving the challenging remaining useful life prediction task in the Predictive Maintenance domain, given the limited amount of training data.

\subsection{Theoretical comparison} \label{sec3.1}

\subsubsection{Comparing the number of parameters}\label{sec3.1.1}

To understand the minimum sample size required to train each candidate model, we first present the mathematical formula for the number of unknown parameters in the considered deep learning models. For the simple RNN described in Section \ref{sec2.2}, supposing that there are $L_{\text{RNN}}$ hidden states, the total number of unknown parameters $\text{Count}_{\text{RNN}}$ is given in Eq.\eqref{count1}. The first term $L_{\text{RNN}}(L_{\text{RNN}}+R)$ in the equation corresponds to the sequential processing by memory cells in Eq.\eqref{SeqDL}, while the second term is related to the output state in Eq.\eqref{SeqDL2}. The number of unknown parameters for LSTM is four times that of RNN, as there are an input gate, an output gate, and a forget gate in addition to the RNN-like memory cell. Gated recurrent units (GRUs) are a type of hidden activation function in recurrent neural networks. GRU is like a long short-term memory (LSTM) with a forget gate, but has fewer parameters than LSTM, as it lacks an output gate. GRU has been used to model speech, music, and language data by maximizing the conditional probability of a target sequence given a source sequence. For the functional MLP in Section \ref{sec2.3} and \ref{sec2.4}, let's denote the length of unknown parameters in the parameter functions $W_{k,r}(\ve{\beta}_{k,r}, t)$ as $Q_{k,r}$ and the number of hidden functional neurons as $L_{\text{FMLP}}$. Then the number of unknown parameters in FMLP $\text{Count}_{\text{FMLP}}$ can be obtained by the formula in Eq.\eqref{count1}. Note that the first term represents the connections between layers and the second term corresponds to the biases in every layer. 

\begin{equation}\label{count1}
\begin{split}
& \text{Count}_{\text{RNN}} = L_{\text{RNN}}(L_{\text{RNN}}+R) + L_{\text{RNN}} = O(L^2_{\text{RNN}})\\
& \text{Count}_{\text{LSTM}} = 4(L_{\text{LSTM}}(L_{\text{LSTM}}+R) + L_{\text{LSTM}})= O(L^2_{\text{LSTM}})\\
& \text{Count}_{\text{GRU}} = 3(L_{\text{GRU}}(L_{\text{GRU}}+R) + L_{\text{GRU}})= O(L^2_{\text{GRU}})\\
& \text{Count}_{\text{FMLP}} = (\sum_{k=1}^{L_{\text{FMLP}}}\sum_{r=1}^R Q_{k,r} + L_{\text{FMLP}}) +(L_{\text{FMLP}}+1)= O(L_{\text{FMLP}})
\end{split}
\end{equation}

Based on Eq.\eqref{count1}, it can be seen that the total number of unknown parameters of RNN, LSTM, and GRU are in a quadratic order of the number of hidden units, while the number of parameters of FMLP is in the same order with $L_{\text{FMLP}}$, the number of functional neurons. This indicates that the minimal sample size required by the sequential learning model is theoretically much larger than FMLP when the underlying mapping $F(\cdot)$ in Eq.\eqref{setting1} and \eqref{setting2} is complex and needs a comparable large number of hidden units in the sequential learning models and FMLP to be well approximated. On the other hand, when $L_{\text{RNN}}$ and $L_{\text{FMLP}}$ are significantly different (i.e., $L_{\text{RNN}} \ll L_{\text{FMLP}}$, or $L_{\text{RNN}} \gg L_{\text{FMLP}}$), the model that can more efficiently capture the temporal information in the covariates requires less number of training samples. As a next step, we study the candidate model's feasibility and efficiency in terms of temporal information capturing under different circumstances. LSTM is the popularly used in many applications, in the rest of the paper, we use LSTM as a representative of the sequential learning models.

\subsubsection{Comparison under different scenarios}\label{sec3.1.2}
In real practice, the observed time series data often contains certain zero-mean noises. That is there exist the additive relationship among the actual observation $Z^{(i,r)}_{j}$, the underlying continuous process $X^{(i,r)}(t)$ that give rises to the time-specific observation, and the random noise $\epsilon_{i,r,j}$, as indicated by Eq.\eqref{setting1.5}. Depending on its mathematical properties, the individual function $X^{(i,r)}(t)$ can be divided into two categories, consisting of smooth functions whose continuous second derivatives exist and non-smooth functions otherwise. Both smooth and non-smooth processes are frequently encountered in real-world applications. Examples of time series data from smooth underlying functions include child growth over time, traffic flow during the day, accumulative number of positive cases over time for a certain pandemic. Continuous processes of the non-smooth nature include the vibration data and acoustic data, of which the rapidly changing dynamics contain the key information to distinguish a given function $X^{(i,r)}(t)$ from other random samples $\{X^{(i,r)}(t)\}_{i^{\prime}\neq i}$. 

According to Section \ref{sec2}, the sequential learning models are more appropriate models for building predictive models with non-smooth time series covariates. This is because the sequential learning models conduct computations on the individual observations at each timestamp and offer a good way of capturing the non-linear dependencies within the highly dynamic time series as well as the complex relationship between the time series and the response. Whereas, the functional predictive models attempt to model the infinite-dimensional continuous curve and therefore essentially rely on certain smoothness assumptions on each underlying process $X^{(i,r)}(t)$ to overcome the curse of dimensionality in the formulation in Eq.\eqref{flm} and \eqref{FN1} \cite{ramsay2006functional}. More specifically, the assumption that the parameter function $W_{k,r}(\ve{\beta}_{k,r},t)$ can be determined by a finite-dimensional vector $\ve{\beta}_{k,r}$, i.e., the assumed model can be trained end-to-end accordingly, only if the underlying process is smooth.

When it comes to smooth time series inputs, the FMLP has advantages over the sequential learning models in general thanks to the basis expansion technique in FDA. The key idea of basis expansion is to set the weight function  $W_{k,r}(\ve{\beta}_{k,r},t)$ as a linear combination of a set of fixed or data-driven basis functions $\{\phi_{k,r,p}(t)\}_{p=1}^{Q_{k,r}}$.
\begin{equation}
W_{k,r}(\ve{\beta}_{k,r},t) = \sum_{p=1}^{Q_{k,r}}\beta_{k,r,p}\phi_{k,r,p}(t). \label{basis1}
\end{equation}
Then the core temporal information processing unit $\int_{t\in \mathcal{T}}W_{k,r}(\ve{\beta}_{k,r},t)X^{(i,r)}(t)dt$ in FMLP becomes
\begin{equation}
\int_{t\in \mathcal{T}}W_{r}(\ve{\beta}_{r},t)X^{(i,r)}(t) = \sum_{p=1}^{Q_{k,r}}\beta_{k,r,p}\int_{t\in \mathcal{T}}\phi_{k,r,p}(t)X^{(i,r)}(t)dt. \label{basis2}
\end{equation}
This means that the infinite-dimensional variation in $X^{(i,r)}(t)$ is transferred to the $Q_{k,r}$ scalar random variables $\{\int_{t\in \mathcal{T}}\phi_{k,r,p}(t)X^{(i,r)}(t)dt\}_{p=1}^{Q_{k,r}}$, the projection scores of $X^{(i,r)}(t)$ onto the pre-defined basis functions. The basis function-based parameter specification in Eq.\eqref{basis1} is supported by theoretical arguments that the infinite continuous stochastic process $X^{(i,r)}(t)$ can be consistently represented by a finite set of basis functions given a certain degree of smoothness \cite{ramsay2006functional,hall2006properties}. The benefit of the formulation in Eq.\eqref{basis2} is two-folds. First, the temporal information in $X^{(i,r)}(t)$ can be more succinctly captured by the integrals, which take less time to compute than RNN, especially when the time series inputs are long time series. Second, it allows us to embed prior domain knowledge about the characteristic of time series features into the temporal predictive model. For instance, if we know there are sparse jumps, spikes, peaks in the time series covariates, the wavelet basis is a good choice to extract meaningful projection scores. On the other hand, the sequential learning models need more complex architectures to capture this useful knowledge. This means that the number of hidden units and accordingly the minimal sample size required in FMLP is in general much less than the sequential learning models for a target level model performance.

\subsection{Comparison of performance on equipment remaining useful life prediction}\label{sec3.2}
Remaining Useful Life (RUL) of equipment or one of its components is defined as the time left until the equipment or component reaches its end of useful life. Accurate RUL prediction is exceptionally beneficial to Predictive Maintenance, and Prognostics and Health Management (PHM). Recently, data-driven solutions that utilize historical sensor and operational data to estimate RUL are gaining popularity. In particular, the RUL prediction problem is usually formulated as a temporal regression problem defined in Eq.\eqref{setting1} and \eqref{setting2}. In this section, we compare the performance of functional MLP (`FMLP'), LSTM, and the traditional multivariate regression models that treat the measurements in time series as features (`MLP' and `SVR') in solving the RUL prediction task with a limited number of training samples in a widely-used benchmark data set called NASA C-MAPSS (Commercial Modular Aero-Propulsion System Simulation) data \cite{saxena2008phm08}. 

%We compare the performance of functional MLP with a variety of state-of-the-art deep learning approaches, including the Convolutional Neural Network model (`CNN') in \cite{babu2016deep}, the Deep Weibull network (`DW-RNN') and the multi-task learning network (`MTL-RNN') in \cite{aggarwal2018two}, the Long Short-Term Memory method (`LSTM') \cite{zheng2017long}, and the bootstrapping based Long Short-Term Memory method (`LSTMBS') \cite{liao2018uncertainty}. As shown by the experimental results, the proposed functional MLP approach significantly outperforms all these alternative methods.

\subsubsection{Background and data pre-processing}
\textit{Background:} C-MAPSS data set consists of simulated 21 sensor readings, 3 operating condition variables for a group of turbofan engines as they running until some critical failures happen. There are four data subsets in C-MAPSS that correspond to scenarios with different numbers of operating conditions and fault modes \cite{saxena2008phm08}. Each subset is divided into the training and testing sets. The training sets contain run-to-failure data where engines are fully observed from an initial healthy state to a failure state. The testing sets consist of prior-to-failure data where engines are observed until a certain time before failure. Table.~\ref{bg} provides a summary for each subset in C-MAPSS, including the number of operating conditions and fault modes, and the number of subjects in the training and testing phase. 

\begin{table}[htbp]
%\vspace{-0.08in}
\caption{Summary of the subsets in C-MAPSS data set}
\vspace{-0.15in}
\begin{center}
\begin{tabular}{c|cccc}
\hline
\hline
\textbf{}& \textbf{FD001}&   \textbf{FD002}&  \textbf{FD003}& \textbf{FD004}\\
\hline
$\#$ of engines in training& 100 & 260 & 100  & 249  \\
$\#$ of engines in testing& 100 & 259 & 100 &  248 \\
$\#$ of operating conditions& 1 & 6 & 1 &   6 \\
$\#$ of fault modes & 1 & 1 & 2 &  2 \\
\hline
\hline
\end{tabular}
\vspace{-0.17in}
\label{bg}
\end{center}
\end{table}

%\textit{4) Normalizing sensors using operating conditions:} 

\begin{figure*}[htbp]
	\centering
	\begin{subfigure}[t]{2.25in}
		\centering
        \caption{Raw sensor data for one randomly selected engine in the training set of FD002.}\label{raw}
        %\vspace{-0.08in}
		\includegraphics[width=4.75cm,height=4.25cm, trim={0 0 20cm 0},clip]{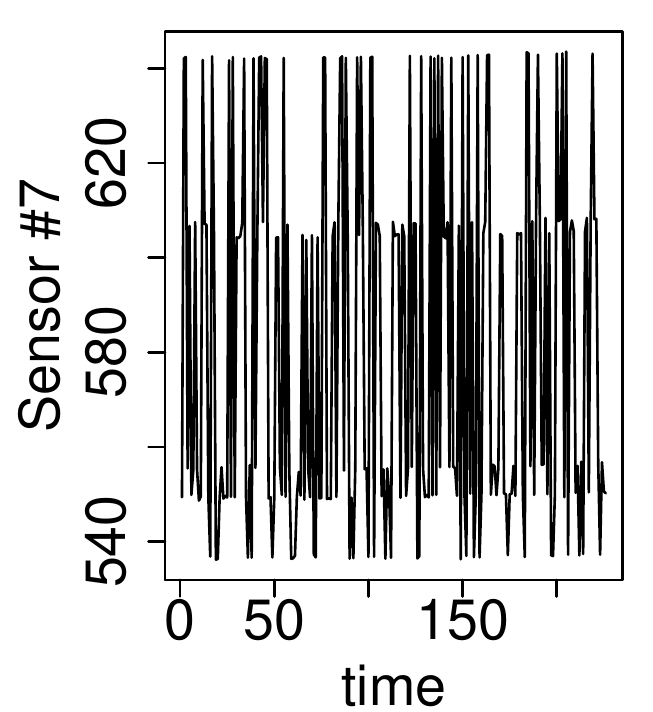}		
	\end{subfigure}
    \quad
	\begin{subfigure}[t]{2.25in}
		\centering
        \caption{Normalized sensor data after removing the effect of operating conditions.}\label{normalized}
        %\vspace{-0.08in}
		\includegraphics[width=4.75cm,height=4.25cm, trim={0 0 20cm 0},clip]{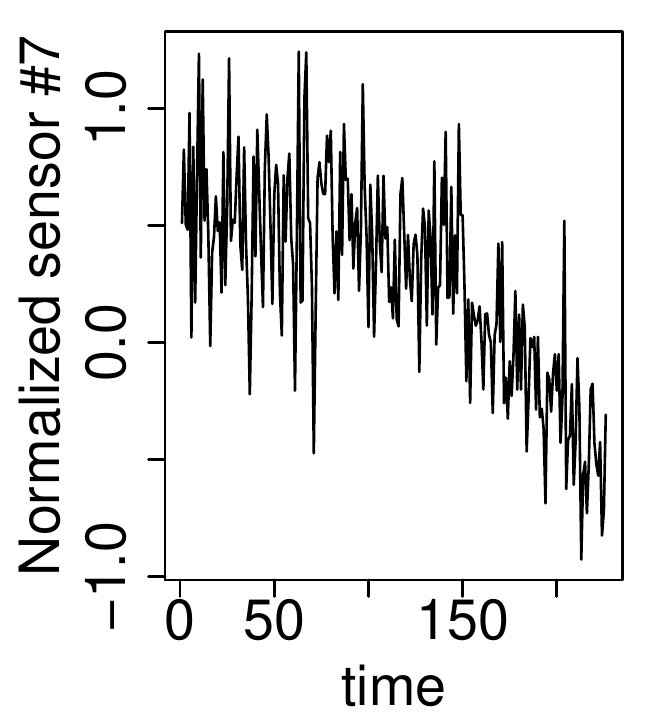}		
	\end{subfigure}
	\vspace{-0.05in}
	\caption{Removing the effect of operating conditions on sensor data.}\label{normalization}
	%\vspace{-0.15in}
\end{figure*}

%\subsubsection{Removing the effect of operating conditions}
\textit{Removing the effect of operating conditions:} For the second and the fourth sub-datasets, there are six operating conditions reflected by the three operating condition variables. To remove the effects of operating conditions, for each of the 21 sensors, we train an MLP model to learn a mapping from the operating condition variables to the sensor variable. In particular, for a given sensor in `FD002', we use data from all the 260 engines in the training set to train the MLP model. The sample size is the summation of the number of observations across the 260 engines. The input data is the three operation condition variables and the output is the considered sensor data. Then we normalize the considered sensor variable by deducting the fitted value of the MLP model from the raw sensor readings. The raw sensor trajectory and normalized trajectory of the second sensor for a randomly selected engine in the training set of FD002 are visualized in Figure.~\ref{normalization}. 

%This normalization step is also considered in \cite{wang2018maintenance, wang2019evaluation}.

%\subsubsection{RUL labeling}
\textit{Window-sliding and RUL labeling:} In the C-MAPSS data set, the engines in the training and testing sets are observed for a different number of time cycles. Moreover, the full sensor data trajectories in the testing sets are blinded for a variety of periods, therefore the true RUL labels are distributed variously. To handle this phenomenon, we propose to use the window sliding technique used in \cite{wu2007neural, tian2012artificial}. Let's denote the smallest number of sensor measurements for the individual engines in data subset $d$ as $\mathcal{M}_d$ for $d=1,...,4$. The values for 
$\mathcal{M}_1, \mathcal{M}_2, \mathcal{M}_3, \mathcal{M}_4$ are $31, 21, 38, 19$ respectively. The functional inputs and RUL labels are generated as follows. For the $d$-th subset, trajectories corresponding to each engine in the training and testing data sets are cut into multiple data instances of length $\mathcal{M}_d$. For instance, the first engine in the training set of FD001 fails at the 144th cycle. A total of 114 training data instances are generated from this engine, with the $c$-th data instance being the sensor measurements between time cycle  $c$ and $c+\mathcal{M}_d-1$. To specify the RUL labels for the 114 data instances of this engine, we adopt the widely-used piece-wise labeling approach in relevant literature \cite{babu2016deep,zheng2017long}. Under the observation that the degradation in the performance is negligible at the beginning period and it starts to degrade linearly at some point $T$, the RUL label is defined as 
\begin{equation}
    \text{RUL}_{c, \text{piecewise}} = \min\{T,  \text{RUL}_{c,\text{linear}}\}.\label{rul_ps}
\end{equation}
Note that in our experiment, we set $T=130$, following the specifications in the prior art \cite{babu2016deep,zheng2017long}.
%The piece-wise labeling approach with $T=130$ is widely utilized in relevant literature \cite{babu2016deep,zheng2017long}. To make the experimental results comparable, we use this RUL labeling approach in our experiment. 

 \begin{figure}[htbp]
     \centering
     \includegraphics[width=4.75cm,height=4.75cm]{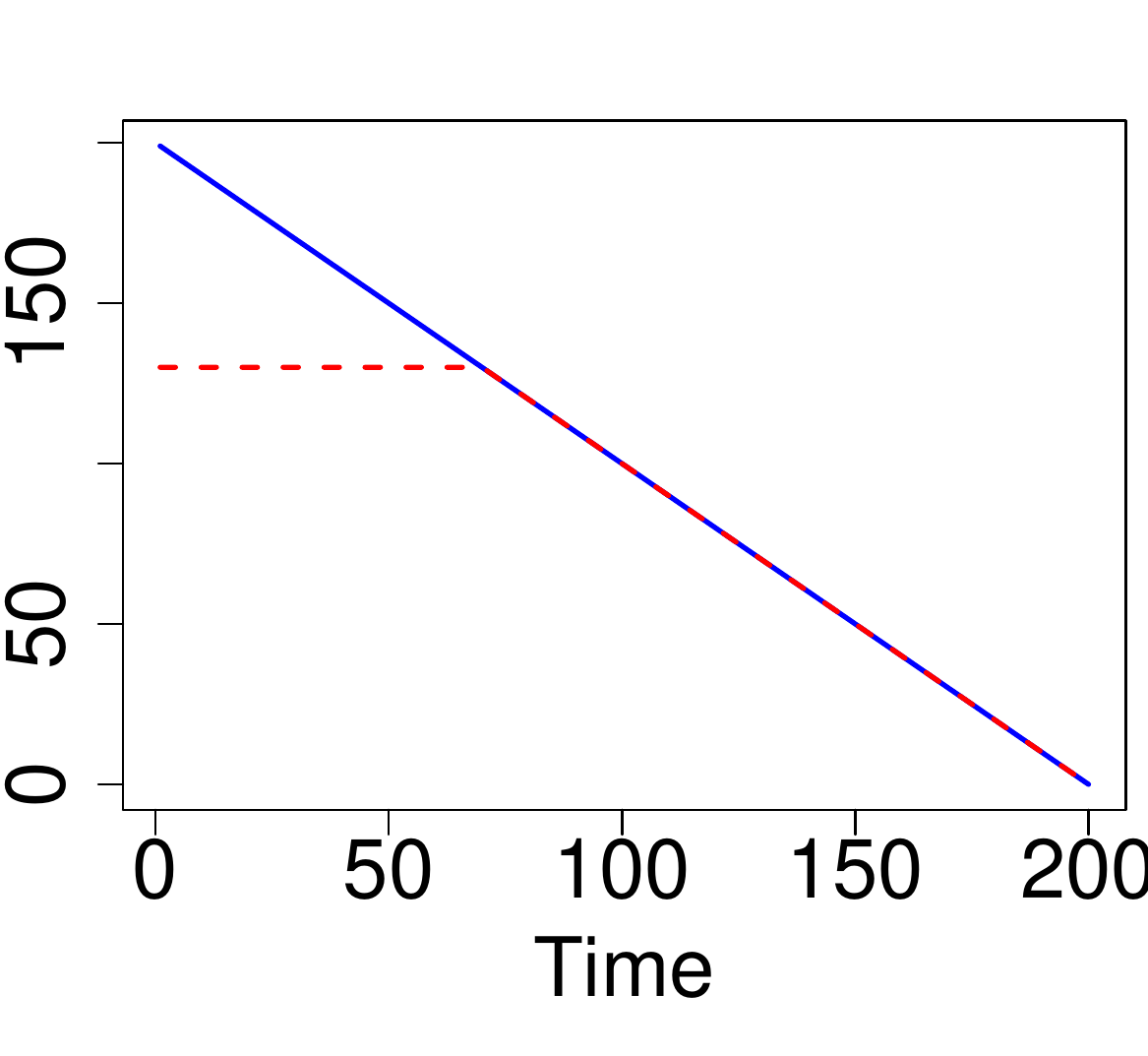}
     \caption{Remaining useful life label for a given engine: the red line represents the piece-wise RUL label capped at $T=130$.}
     \label{fig:my_label}
 \end{figure}

\subsubsection{Implementations and results}

\textit{Implementation of FMLP:} Following the implementation of LSTM in \cite{zheng2017long}, we use the Min-Max normalization to scale the individual sensor sequences to the range of $[0,1]$. The specific mathematical formula can be found in \cite{zheng2017long, wang2019remaining}. A FMLP with a two-layered architecture is deployed to learn the mapping from 21 sensors to the RUL label. There are four functional neurons (i.e., K=4) on the first layer and two numerical neurons on the second layer in the FMLP. The activation function on both layers are the standard logistic function, i.e., $U_{k}(u) = \frac{1}{1+e^{-u}}$. To better deal with the complex sensor data, we propose to specify data-driven weight functions by calculating the eigenfunctions from data. Let the estimated eigenfunction from the $N$ samples of the $r$-th sensor be $\hat{\phi}_{r,p}(t)$. The weight functions are then specified as
\begin{equation}
\label{exp1_d}
W_{k,r}(\ve{\beta}_{k,r},t) = \sum_{p=1}^{P_{k,r}} \beta_{k,r,p}, \hat{\phi}_{r,p}(t),
\end{equation}
where $P_{k,r}$ is selected from the regularly-used fraction of variance explained (FVE) with a $80\%$ cutoff. In practice, there are four commonly-used cutoff values, i.e., $80\%$, $90\%$, $95\%$, and $99\%$. In this experiment, we choose the smallest rule-of-thumb value. This is because choosing a smaller FVE helps retaining the key smooth patterns in sensor data and removing the random noises shown in Figure \ref{normalization}.

\textit{Evaluation metrics:} We evaluate the performance of functional MLP with the same evaluation strategy used in \cite{babu2016deep,zheng2017long}. Suppose that there are $N$ subjects in the testing set, and the true RUL since the last observation of engine $i$ is $\text{RUL}_{i, true}$ and the estimated RUL is $\text{RUL}_{i, est}$. The root mean squared error (RMSE) calculated from the $N$ engines is defined as
\begin{equation}
    \text{RMSE} = \sqrt{\frac{1}{N}\sum_{i=1}^N (\text{RUL}_{i, est}-\text{RUL}_{i, true})^2}.
\end{equation}

\textit{Results:}
The RMSE of FMLP together with the results of LSTM and the multivariate regression models including the Support vector regression (`SVR') and the multilayer perceptron (`MLP') from previous literature are summarized in Tables \ref{tab1d}. For all the four subsets, functional MLP significantly outperforms the baseline methods in terms of RMSE. The average improvement over LSTM \cite{zheng2017long} is $26.89\%$. For industrial equipment like turbofan engines, the sensor signals over time are often correlated with the smooth degradation process and thus can be assumed to be generated by smooth continuous functions. The experimental results in Tables \ref{tab1d} numerically justify our discussion about the advantage of FMLP over the state-of-art models in handing smooth time series covariates in Section \ref{sec3.1}.

\begin{table}[h]
%\vspace{-0.05in}
\caption{RMSE comparison on C-MAPSS data and improvement (`IMP') of functional MLP over LSTM \cite{zheng2017long}.}
\begin{center}
 \vspace{-0.2in}
\begin{tabular}{c|cccc}
\hline
\hline
\textbf{Model}& \textbf{FD001}&   \textbf{FD002}&  \textbf{FD003}& \textbf{FD004}\\
\hline
MLP\cite{babu2016deep}& 37.56 & 80.03 & 37.39  & 77.37  \\
SVR\cite{babu2016deep} & 20.96 & 42.00 & 21.05 &  45.35 \\
%RVR\cite{babu2016deep} & 23.80 & 31.30& 22.37 &   34.34 \\
%CNN\cite{babu2016deep} & 18.45 & 30.29 & 19.82 &   29.16 \\
%DW-RNN\cite{aggarwal2018two} & 22.52 & 25.90 & 18.75 &   24.44 \\
%MTL-RNN\cite{aggarwal2018two} & 21.47 & 25.78 & 17.98 &   22.82 \\
%LSTMBS\cite{liao2018uncertainty} & 14.89 & 26.86 & 15.11 &   27.11 \\
LSTM\cite{zheng2017long} & 16.14 & 24.49& 16.18 &   28.17 \\
FMLP& \textbf{13.36} & \textbf{16.62} & \textbf{12.74}  & \textbf{17.76}  \\
\hline
\hline
IMP& $17.22\%$ & $32.14\%$ & $21.26\%$  & $36.95\%$  \\
\hline
\hline
\multicolumn{5}{l}{\footnotesize * IMP w.r.t LSTM is $(\text{RMSE}_{\text{LSTM}}-\text{RMSE}_{\text{LSTM}})/\text{RMSE}_{\text{LSTM}}$}\\
\end{tabular}
\label{tab1d}
\vspace{-0.1in}
\end{center}
\end{table}

\section{Scarce data with sparse time series features}\label{sec4}
In this section, we consider circumstances where the time series inputs are sparsely and irregularly observed over the time domain. The sparse functional MLP (SFMLP) in Section \ref{sec2.4.1} and the SFMLP equipped with multivariate FPCA in Section \ref{sec2.4.2} are temporal predictive models that are specially designed to handle this type of scenario. The sequential learning models are incapable of directly utilizing the sparsely evaluated temporal information for proper model building. They require to preliminarily fill in the gaps in the raw sparse data based on certain interpolation techniques. The performance of the sequential learning models heavily relies on the accuracy of interpolation. A description of this interpolation plus sequential learning approach is provided in the first part of this Section. Next, we conduct three numerical experiments to demonstrate the superior performance of the proposed sparse FMLPs for sparse data scenarios.

\subsection{Sequential learning models under sparse data} 
As discussed in Section \ref{sec2.2}, the sequential deep learning models essentially require the time series covariates to be densely observed at an equally-spaced time grid. Under sparse data cases, certain data interpolation techniques such as the cubic spline interpolation are often deployed to get data readings of the same interval. The interpolated data are then fed into sequential learning models to build temporal predictive models. The performance of the sequential learning models heavily relies on the accuracy of interpolation.

\begin{figure}[h]
\centering\includegraphics[width=7.0cm,height=4.0cm]{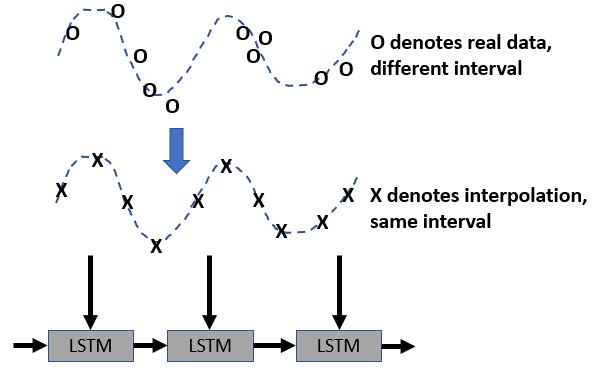}
\vspace{-0.1in}
\caption{Data pre-processing in sequential learning models.}
\label{lstm}
\end{figure}

\begin{figure}[htbp]
\centering
\begin{tabular}{cc}
  \includegraphics[width=45.7mm, height=37.2mm]{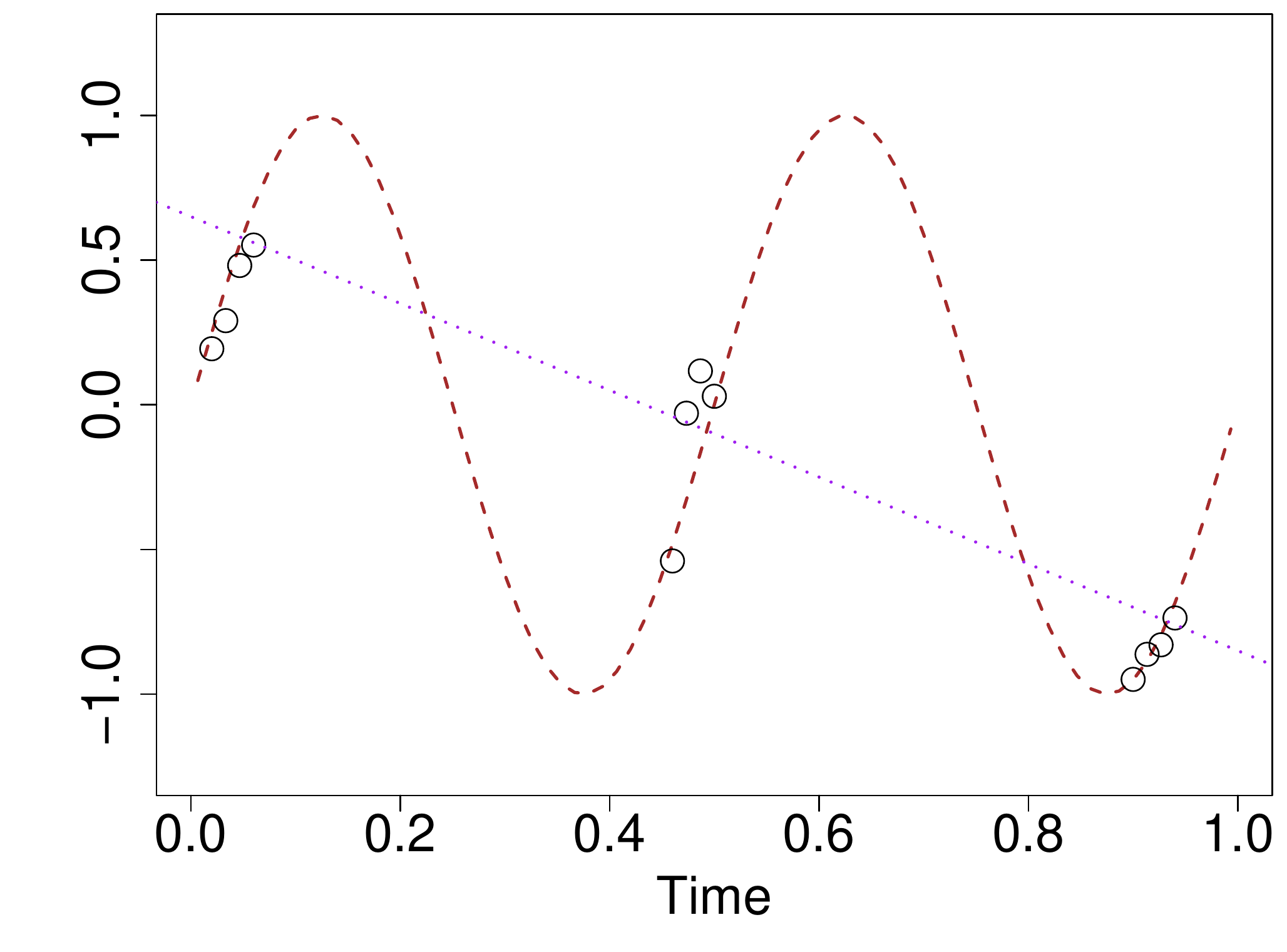} & \includegraphics[width=45.7mm, height=37.2mm]{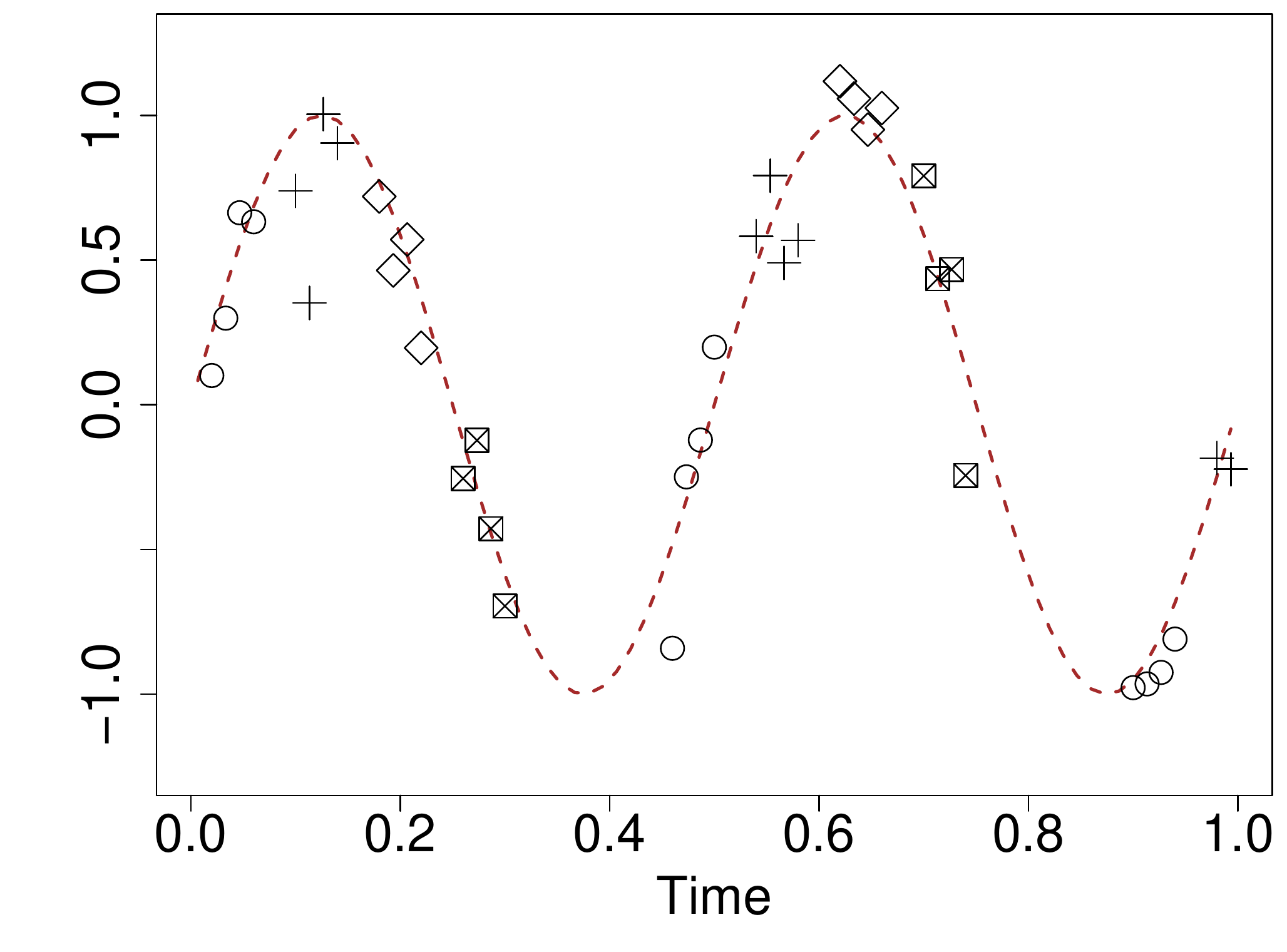} \\
\end{tabular}
\vspace{-0.2in}
\caption{Left: An example when doing interpolation based on a single subject is problematic. Brown line is the underlying true curve, black dots are the observations to be used, and purple lines is the biased pre-smoothing results. Right: Visualization of the benefit of performing interpolate using combined the observations from different subjects.}
\label{int}
\end{figure}

The conventional way of performing interpolation is to separately apply techniques such as cubic B-spline and Gaussian process regression on the sparse observations from each subject. This common practice is illustrated in Figure ~\ref{lstm}. However, the recovered curves using data from the individual curves alone are often not consistent estimates for the true underlying curves, due to the limited amount of information available per curve. For instance, as shown by the left figure in Figure ~\ref{int}, the interpolated curves are significantly dispersed from the actual $\sin$-shaped random functions with two full cycles. As a consequence, the sequential learning models built based upon the biased input data are not reliable. 

Observing the benefit of jointly using time series across samples to estimate or model the temporal variations within time series data in functional data analysis (i.e., the right figure in Figure ~\ref{int}), in this paper, we also consider the functional data approach PACE in Eq.\eqref{sparse2} and \eqref{sparse5} as an interpolation method for the sequential learning models. As shown in the numerical experiments in Section \ref{sec5.2}, the performance of LSTM when the recovered data are generated by PACE is significantly better than the other interpolations for the considered problems. In general, FDA type of models is valid when the time series of different subjects can be considered as random samples from an underlying random process, which is a reasonable assumption under many real-world use cases. Therefore, we recommend using FDA-type modelings as interpolation when dealing with sparse time series.

\subsection{Performance comparison in numerical experiments}\label{sec5.2}
In this section, we conduct three numerical studies, including classification of synthetic curves, prediction of patient's survival beyond a given period, and prediction of engine's remaining time to failure. We compare our sparse functional MLP (`sparse FMLP' or `SFMLP') with the `LSTM' as the baseline methods. Our proposed method outperforms LSTM in all three numerical studies, where the input variables are observed multiple times for each subject, and the observing times are irregularly spaced and are not shared across subjects.
\subsubsection{Curve classification on simulated data}
\label{sec4.1}
In this subsection, we conduct a numerical experiment on synthetic data, with the objective of showing the benefit of modeling repeated data over time in the functional fashion as well as the validity of the proposed sparse functional MLP. Specifically, we consider a curve classification problem. For each subject, we have a variable of interest measured at multiple random times within time range $\mathcal{T}$. Each subject has an associated group label. The problem is to build a model to predict the group label using the repeatedly observed feature within time window $\mathcal{T}$.

\begin{figure}[h]
	\centering
	\begin{subfigure}[t]{3.5in}
		\centering
        \caption{Two randomly selected subjects in group 1.}\label{exp1_1}
        \vspace{-0.1in}
		\includegraphics[width=7.5cm,height=3.2cm]{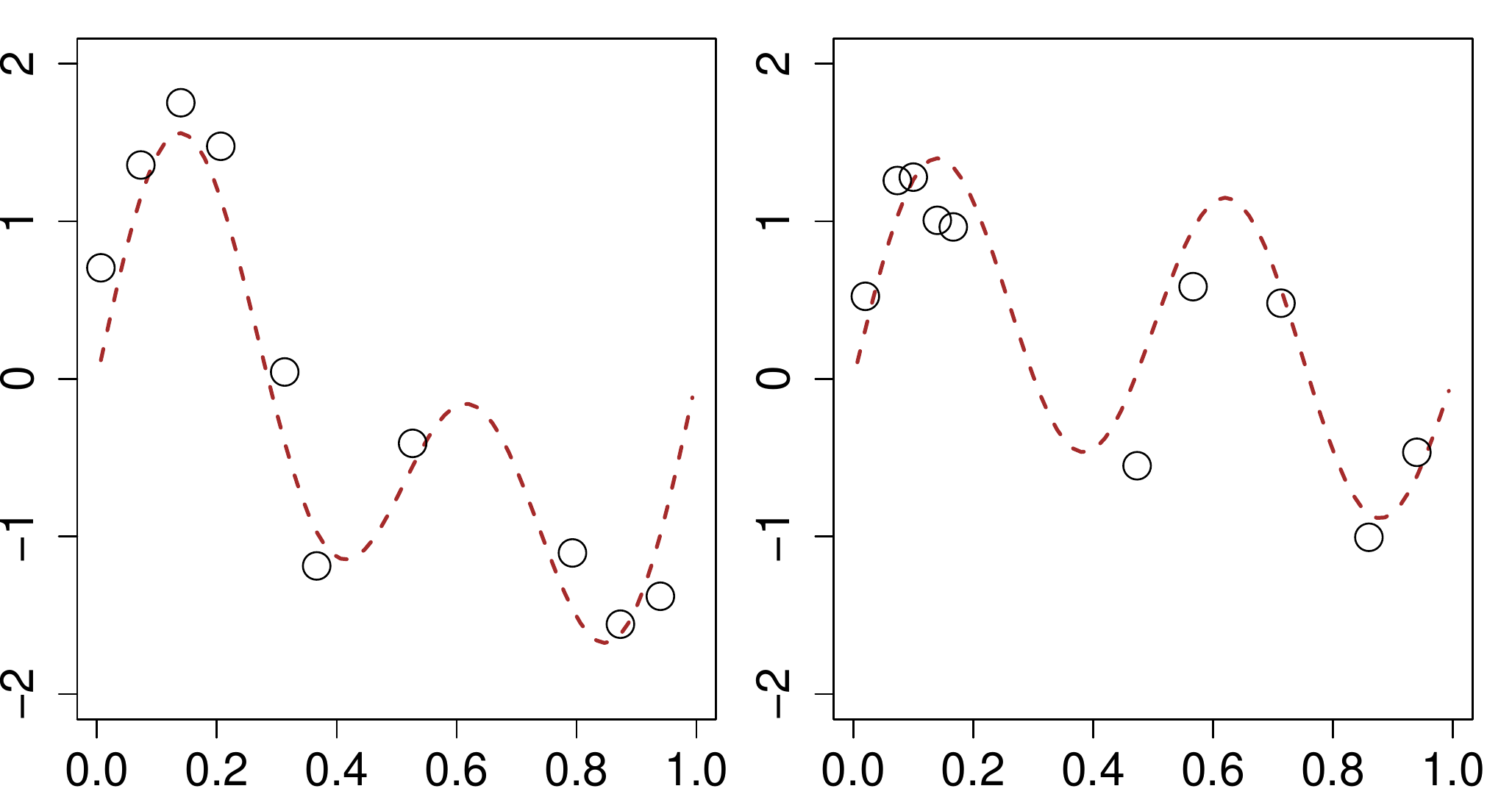}		
	\end{subfigure}
	\par\bigskip
	\begin{subfigure}[t]{3.5in}
		\centering
        \caption{Two randomly selected subjects in group 2.}\label{exp1_2}
        \vspace{-0.1in}
		\includegraphics[width=7.5cm,height=3.2cm]{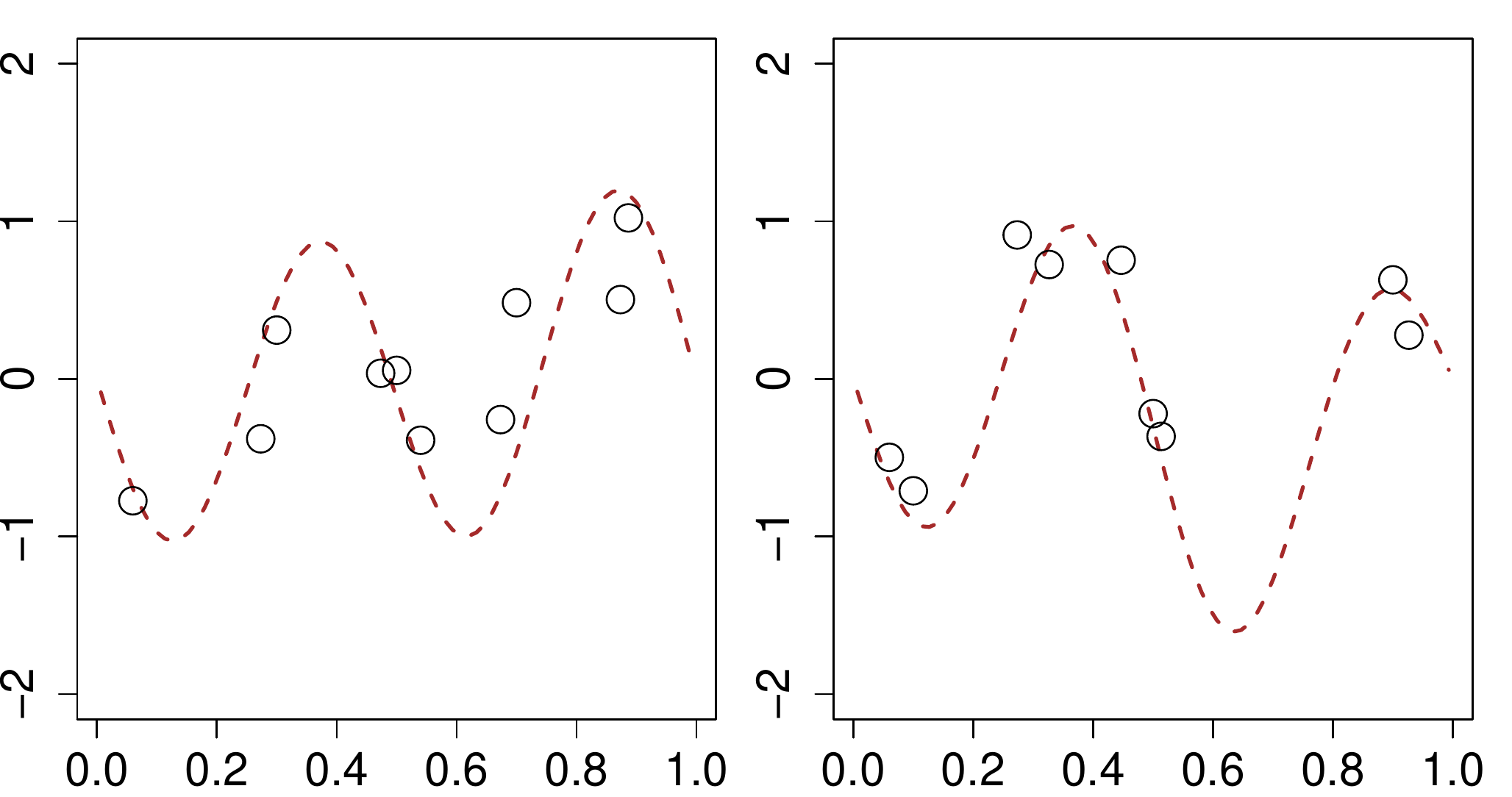}		
	\end{subfigure}
	\caption{Visualizations of simulated data in Section \ref{sec4.1}.}\label{exp111}
\end{figure}

The synthetic curves and labels are generated as follows. There are two distinct groups, i.e., $g=1,2$. The $j$-th observation of the $i$-th subject in group $g$ is denoted as $Z_{g,i,j}$, which is generated using $Z^{(g,i)}_{j}=X^{(g,i)}(T^{(g,i)}_{j}) + \epsilon_{g,i,j}$ Eq.\eqref{setting1.5} and the Karhunen-Lo{\`e}ve expansion $X^{(g,i)}(T^{(g,i)}_{j})=\mu_g(T^{(g,i)}_{j})+\sum_{p=1}^{\infty}\xi_{g,i,p}\phi_{p}(T^{(g,i)}_{j})$\cite{ramsay2006functional}, with $\xi_{g,i,p}\sim N(0,\lambda_p)$, for $i=1,...,N_g$, $j=1,...,M_{g,i}$. Without loss of generality, we set time window $\mathcal{T}=[0,1]$. The number of subjects in both groups are $N_1=N_2=300$. The number of observation on each curve is 10, i.e., $M_{g,i}=M=10$. Given $M_{g,i}$, the observing times are i.i.d samples from the Uniform distribution within [0,1]. The $p$-th eigenfunction $\phi_{p}(t)$ equals $\sqrt{2}\sin{(p\pi t)}$ for $p=1,...,\infty$. The first four eigenvalues are $(0.1, 0.045, 0.01, 0.001)$ and $\lambda_p=0$ for $p>4$. The mean functions are $\mu_1(t)=\sin{(4\pi t)}$ and $\mu_2(t)=-\sin{(4\pi t)}$. The standard deviation of the i.i.d random errors is 0.3. Two randomly selected subjects from each of the two groups are visualized in Figure.~\ref{exp1_1} and Figure.~\ref{exp1_2}. In each plot, the true curve $X_{g,i}(t)$ is the brown dashed line and the observations to be modeled are the black dots. We can see that the data on each curve is sparse, and the big gaps between observations prohibit the pre-smoothing step in LSTM and the dense functional MLP to successfully recover the two-cycle sine functions using the limited amount of data available. Our proposed sparse functional MLP is specifically designed to handle this kind of scenarios.

Cubic spline, the Gaussian process regression, and functional data analysis based PACE interpolations are used in this experiment to get data readings of the same interval. We then feed these interpolation data readings into an LSTM network. This LSTM uses two layers LSTM (32, 64) and one layer of neural network (8). Note that grid search is used to tune the hyperparameters in LSTM. Figure.~\ref{exp1_44} shows the achieved interpolation at 100 points ($M=100$) within the period for one randomly selected subject in group 1. It can be seen that cubic splines and the Gaussian process regression produce biased estimates of the true curve, while the result of PACE is consistent. The same observation is shown by the RMSE for all the 300 curves in group 1 in the first table in Table.~\ref{exp_tab1}. Given the simulation result, in real practice, we recommend trying PACE as an alternative curve fitting method when the measurements in each time series are sparse and irregular. The leave-one-out cross validation results are given by the first three rows of Table.~\ref{exp_tab1}.

\begin{figure}[h]
\centering\includegraphics[width=9.5cm,height=4.5cm]{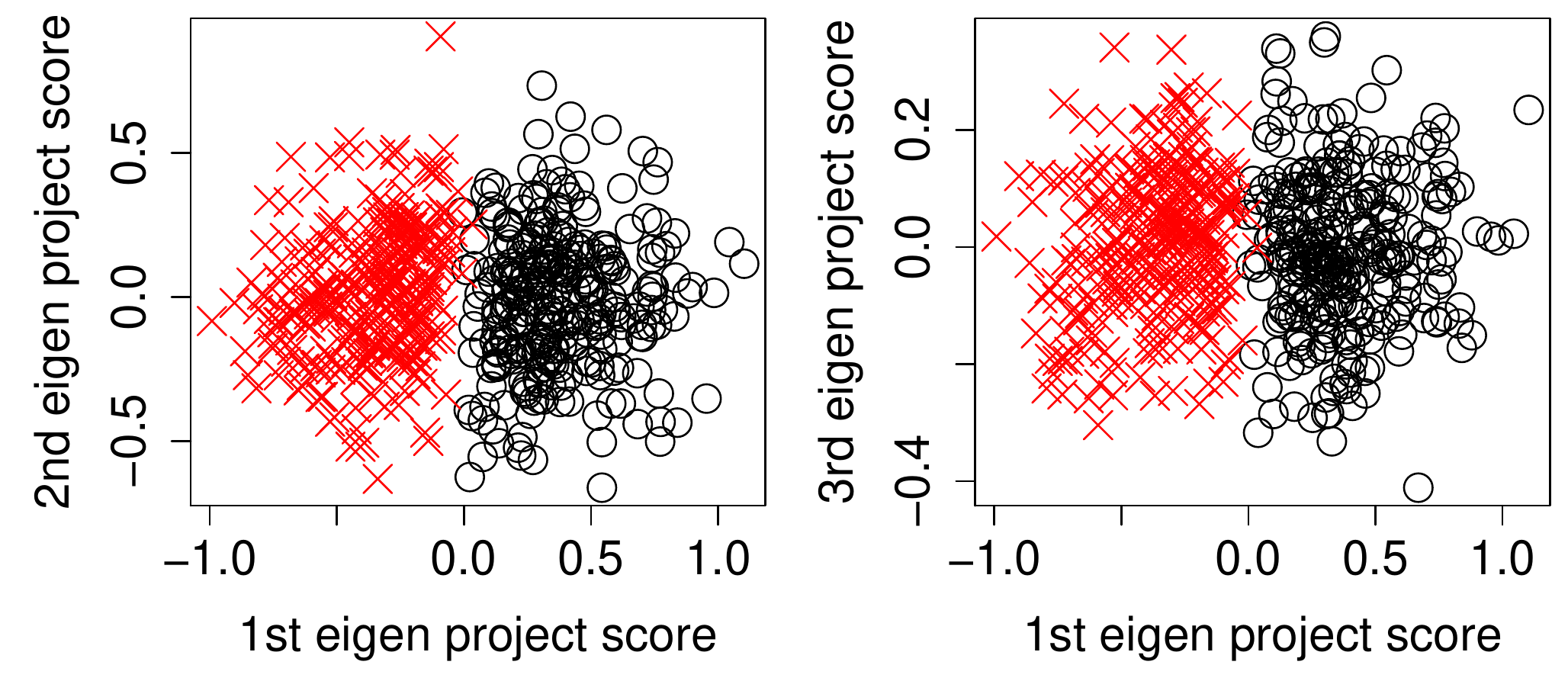}
\vspace{-0.1in}
\caption{Extracted features from sparse FMLP($P=3$). Black dots represent group 1 and red dots represent group 2.}
\label{exp1_4}
\end{figure}

\begin{figure}[h]
\centering\includegraphics[width=4.5cm]{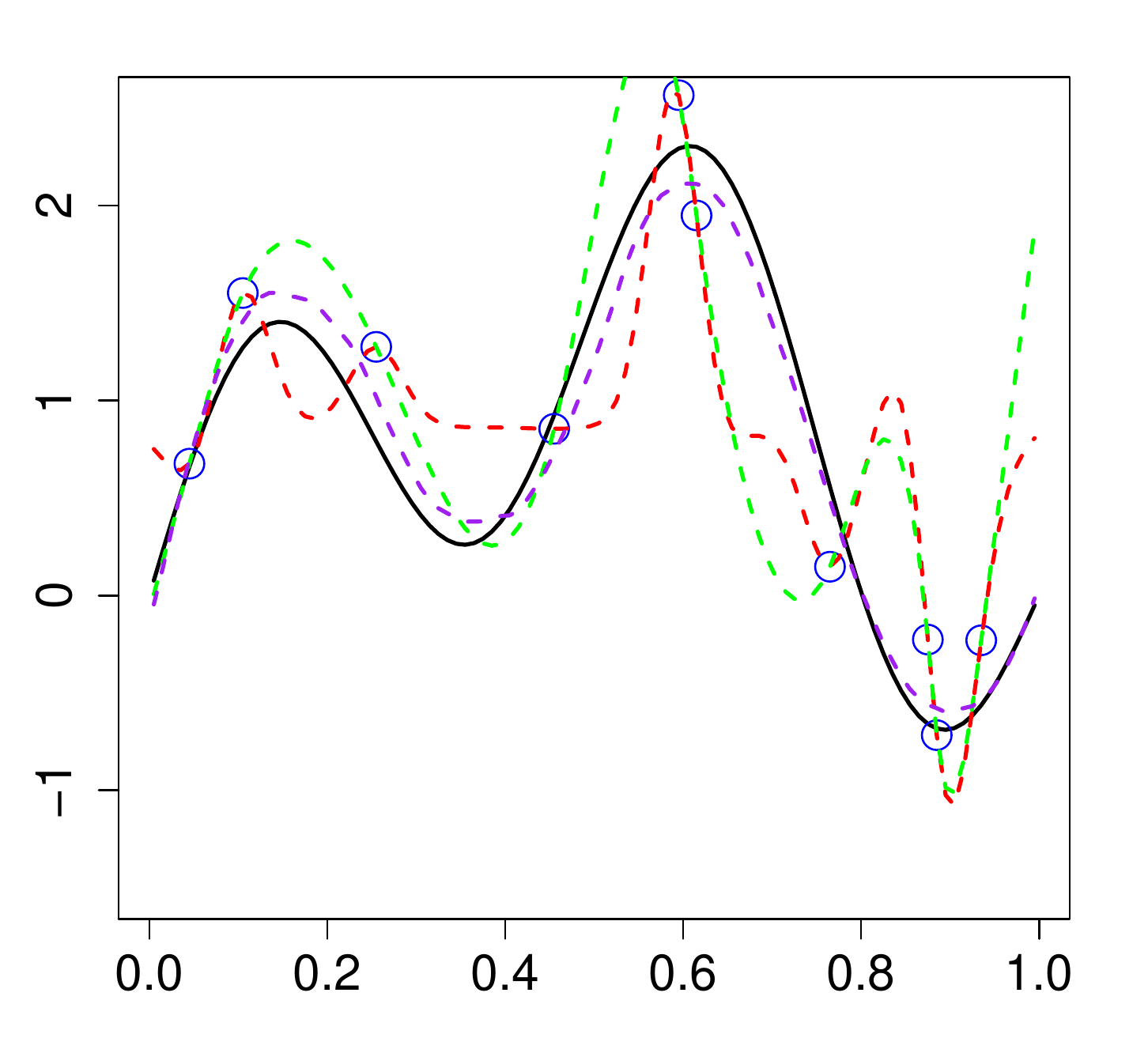}
\vspace{-0.1in}
\caption{Comparison of interpolation results for a randomly selected sample. The blue dots are the sparse observations available in the data st. The black line is the ground truth. The red, green, and purple lines respectively corresponding to the Gaussian process regression, cubic B-spline and PACE. }
\label{exp1_44}
\end{figure}

To implement our proposed sparse FMLP, we first estimate all the required components in the proposed sparse FMLP. The first three dimensions of the achieved eigen projection scores, i.e., $E[\eta_{g, i,1}|\mathbf{Z}_{g,i}], E[\eta_{g,i,2}|\mathbf{Z}_{g,i}]$, and $E[\eta_{g,i,3}|\mathbf{Z}_{g,i}]$ for $g=1,2$ ;$i=1,...,N_g$, are visualized in Figure.~\ref{exp1_4}. In the plots, two different types of dots correspond to the two different groups. As shown by the plots, the extracted eigen projection scores better distinguishes the two evolution curves $\sin(4\pi t)$ and $-\sin(4\pi t)$. We specify the weight function of the $k$-th functional neuron as $W_{k}(\ve{\beta}_{k}, t) = \sum_{p=1}^{P} \beta_{k, p} \hat{\phi}_p(t)$, with $\hat{\phi}_p(t)$ being the $p$-th estimated eigenfunction. The architecture of the sparse functional MLP is that there are 4 functional neurons in the first layer followed by a layer with two numerical neurons. The activation functions in both layers are logistic function. The leave-one-out cross-validation results are given by the last two rows of Table.~\ref{exp_tab1}.

Based on the two tables in Table.~\ref{exp_tab1}, we have the following observations. First, the functional data based interpolation that jointly consider the data from all samples significantly outperforms the alternative techniques. Second, for this example, the performance of our sparse functional MLP under $P=2$ and $P=3$ is comparable, with the result under $P=3$ being slightly worse. This makes sense because the first two and three components contain a comparable amount of information in the curves, according to our setting (i.e., eigenvalues are (0.1, 0.045, 0.01, 0.001)). Also, the reason why $P=3$ slightly worse is that the third component with a very small eigenvalue and can be considered as additional noises introduced to the first two components. Third, the performances of LSTM with different interpolations are similar, although the fitting performance of PACE is significantly better than the other approaches. We think this is because the interpolation within each group might be biased in a similar way for all subjects, such that the two groups of curves are still distinguishable. Last but not least, the performance of LSTM with PACE and the proposed sparse FMLP is comparable. We think this is because the distinction between the two groups of data is large and different models tend to have similar performances.

%LSTM has comparable performance in this example, however, its performance highly depends on the choices of curve recovery methods. And the recovered curves are not theoretically consistent due to the sparseness within each curve. The inconsistency is visually demonstrated by the discrepancy among curves in Figure.~\ref{exp111} and Figure.~\ref{lstm12}. 

%For the simulated scenario, our proposed sparse functional MLP outperforms LSTM in terms of accuracy. 

\begin{table}[h]
\caption{The interpolation accuracy of different methods for time series in group 1 and leave-one-out cross-validation results for synthetic curve classification task in Section \ref{sec4.1}.}
\begin{center}
\begin{tabular}{cc}
\hline
\hline
\textbf{Interpolation method}&  \textbf{RMSE}\\
\hline
Cubic spline& 0.562   \\
Gaussian process regression& 0.440  \\
PACE & \textbf{0.169}\\
\hline
\hline
%\multicolumn{4}{l}{$^{\mathrm{a}}$Sample of a Table footnote.}
\end{tabular}

\bigskip
\begin{tabular}{c|c|c|c|c}
\hline
\hline
\textbf{Model}& \textbf{Spec}&   \textbf{$\#$ of samples}&  \textbf{Architecture}& \textbf{Accuracy}\\
\hline
%LSTM_{\text{Spline}}& $M=10$  &600   & L(32,64)N(8) &   $99.2\%$  \\
$\text{LSTM}_{\text{Spline}}$& $M=100$ & 600  & L(32,64)N(8) &   $99.4\%$  \\
$\text{LSTM}_{\text{GP}}$& $M=100$ & 600  & L(32,64)N(8) &   \textbf{99.5}$\%$  \\
$\text{LSTM}_{\text{PACE}}$ & $M=100$ & 600  & L(32,64)N(8) &   $99.3\%$  \\
Sparse FMLP& $P=2$& 600& F(4)N(2)  & \textbf{99.5}$\%$   \\
%\hline
Sparse FMLP& $P=3$& 600& F(4)N(2)  & $99.3\%$   \\
\hline
\hline
%\multicolumn{4}{l}{$^{\mathrm{a}}$Sample of a Table footnote.}
\end{tabular}

\label{exp_tab1}
\end{center}
\end{table}

\subsubsection{Prediction of PBC patient's long-term survival}
In this section, we consider the problem of predicting the long-term survival of patients with primary biliary cirrhosis (PBC) using their serum bilirubin measurements (in mg/dl) at the beginning period of the study. This enables the doctors to get early warnings and to take corresponding actions to increase patient's survival possibility. The PBC data set we used are results of a Mayo Clinic trial from 1974 to 1984. This data set is publicly available in a $R$ package called `survival' and has been investigated by numerous researchers including \cite{muller2005functional}. 

\begin{figure}[htbp]
	\centering
	\begin{subfigure}[t]{1.75in}
		\centering
        \caption{Histogram of number of observation per patient within the first 910 days in the PBC study.}\label{fig1}
        \vspace{-0.1in}
		\includegraphics[width=4.55cm,height=4.2cm]{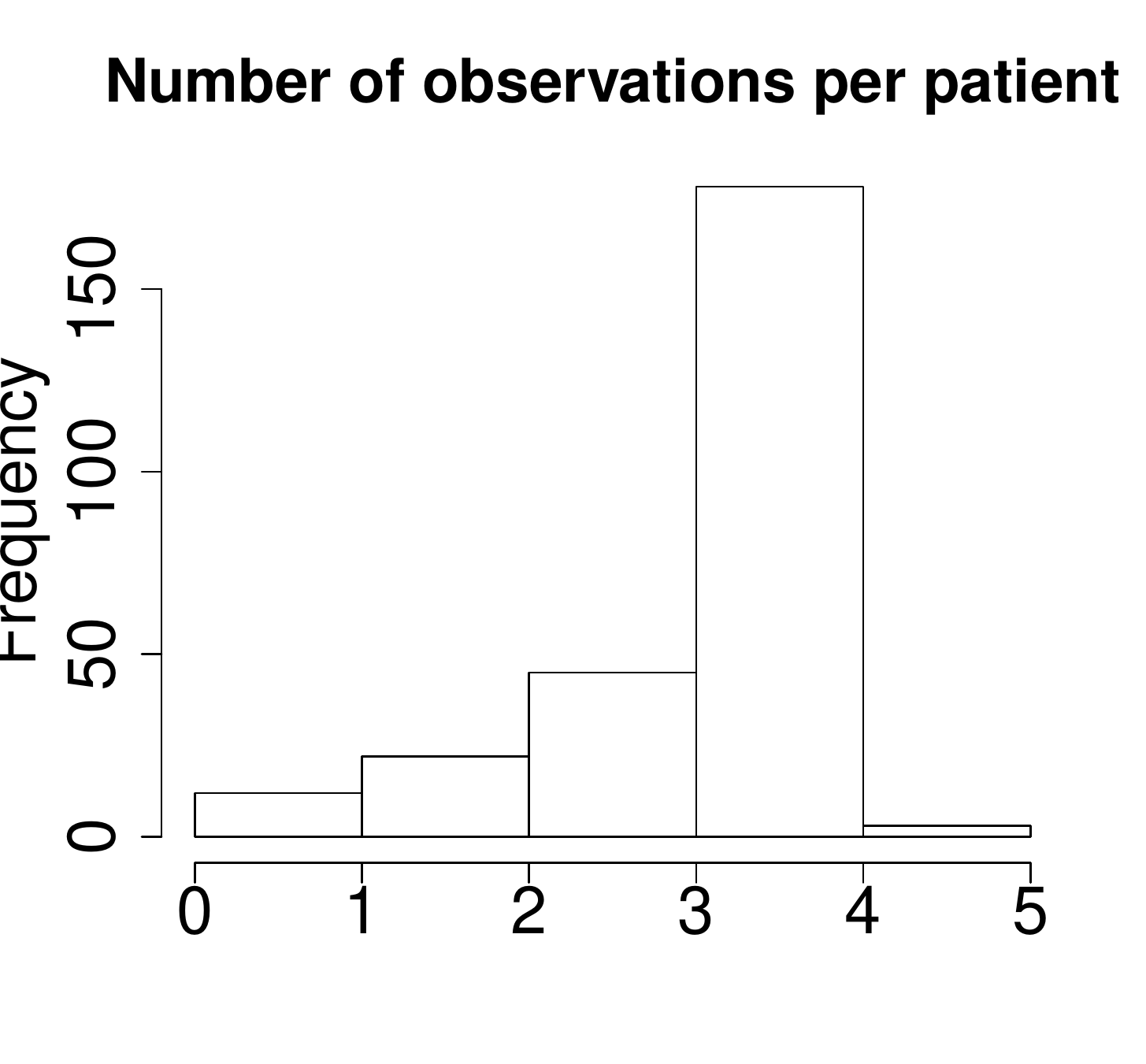}		
	\end{subfigure}
    \quad
	\begin{subfigure}[t]{1.75in}
		\centering
        \caption{Estimated eigenfunctions with first two highest eigenvalues from sparse FMLP.}\label{fig3}
        \vspace{-0.1in}
		\includegraphics[width=4.2cm,height=3.75cm]{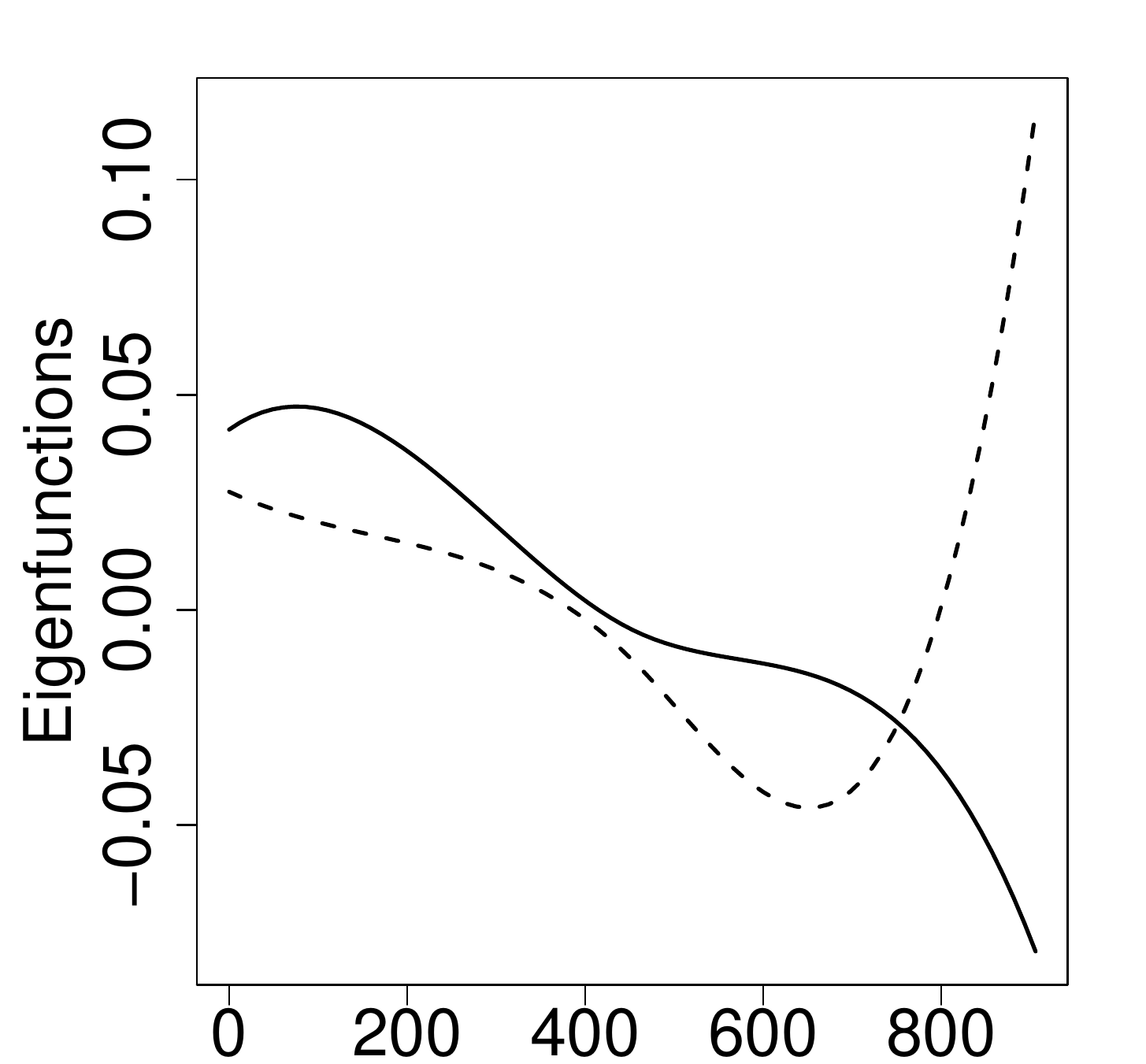}		
	\end{subfigure}
	\caption{Some results for the PBC long-term survival prediction study.}
\end{figure}

\begin{figure}[htbp]
\centering\includegraphics[width=9.75cm,height=3.95cm,trim={0 250.5 0 0},clip]{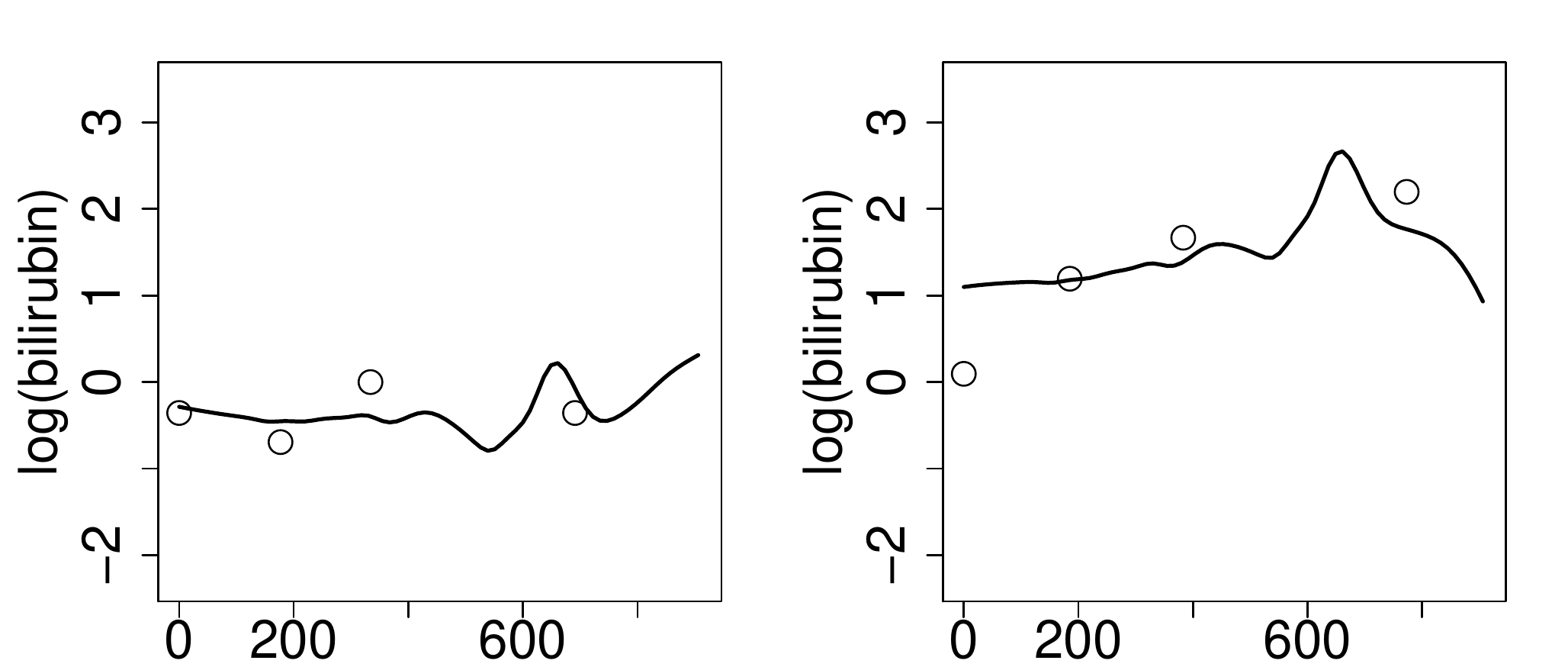}
\vspace{-0.01in}
\caption{The raw data and the recovered log(bilirubin) curve using sparse FMLP for two randomly selected patients. }
\label{fig4}
\end{figure}

We consider the following problem setting. We use the patient's bilirubin measurements, which is known to be an important indicator of the presence of chronic liver cirrhosis, within the first 910 days of the study to predict whether the patients survive beyond 10 years after entering the study. There are 260 patients included in the analysis, with $84$ died between 910 and 3650 days and 176 being alive after 10 years. Following \cite{muller2005functional}, the bilirubin measurements are log-transformed. The number of bilirubin measurements per patient within the 910 days ranges from 1 to 5, with histogram given in Figure.~\ref{fig1}. The sparse observations of two randomly selected patients are plotted as the black dots in Figure.~\ref{fig4}. Given these two plots, it can be seen that the functional data is sparse, and the number of observations and the observing time are not the same across patients. 

Our proposed sparse functional MLP is implemented in the same fashion as described in the previous numerical experiment. The number of projection is chosen through the leave-one-curve-out cross-validation approach described in \cite{yao2005functional}. The selected $\hat{P}=2$. The estimated eigen-functions using the restricted maximum likelihood estimate method in \cite{peng2009geometric}, denoted as $\hat{\phi}_{p}(t)$ for $p=1,2$, are given in Figure.~\ref{fig3}. The best predicted curves using $\tilde{X}_i(t)=\sum_{p=1}^{\hat{P}} \hat{E}[\eta_{i,p}|\mathbf{Z}_{i}]\hat{\phi}_{p}(t)$ for two randomly selected patients are given in  Figure.~\ref{fig4}. In each of the two plots in Figure.~\ref{fig4}, the observed dots are closely distributed around the predicted log(bilirubin) curve $\tilde{X}_{i}(t)$ with some random errors, which visually justified the validity of our curve approximator in Eq.\eqref{sparse55} and \eqref{sparse6}. The architecture of the sparse functional MLP we used is that there are four functional neurons on the first layer, i.e., $K=4$ in Eq. \eqref{sparse55} and \eqref{sparse6}, followed by another layer of two numerical neurons. The activation functions in both layers are logistic function. The weight function $V_{k}(\ve{\beta}_{k}, t)$ for the $k$-th functional neuron in the first layer 

\begin{equation}
\label{exp1}
W_{k}(\ve{\beta}_{k}, t) = \sum_{p=1}^{\hat{P}} \beta_{k, p} \hat{\phi}_p(t),
\end{equation}
with $\hat{\phi}_p(t)$ being the $p$-th estimated eigenfunction. The leave-one-out cross-validation results using the specifications mentioned above are given in Table.~\ref{tab1}. The overall accuracy is 73.08$\%$. We also implemented the cubic B-spline interpolation plus LSTM strategy described in the previous subsection. The accuracy of LSTM with $M=910$ is 67$\%$, which is lower than our proposed method.

\begin{table}[h]
\caption{Leave-one-out cross-validation results for PBC long-term survival prediction using sparse FMLP}
\vspace{-0.2in}
\begin{center}
\begin{tabular}{|c|c|c|}
\hline
\textbf{}&\multicolumn{2}{|c|}{\textbf{True}} \\
\cline{2-3} 
\textbf{Classified} & \textbf{\textit{Survived}}& \textbf{\textit{Died}}\\
\hline
\textbf{\textit{Survived}}& 151& 25     \\
\hline
\textbf{\textit{Died}}& 45&   39\\
\hline
%\multicolumn{4}{l}{$^{\mathrm{a}}$Sample of a Table footnote.}
\end{tabular}
\label{tab1}
\end{center}
\end{table}

\subsubsection{Predicting remaining useful life for aircraft engines}
In this subsection, we re-consider the RUL prediction problem in Section \ref{sec3}. The C-MAPSS data set is a simulated data set without any irregularity and missing values. To mimic real scenarios where there are usually a certain level of irregularity in time series trajectories, we sparsify the data set by randomly keep a certain percentage of the raw data ($30\%$,  $50\%$ or $100\%$) for each of the engines in the training data. Note that the sampled timestamps are different for different variables and subject. The last observation of each engine is always kept to indicate the failure time.

\begin{figure*}[htbp]
	\centering
	\includegraphics[width=14.75cm]{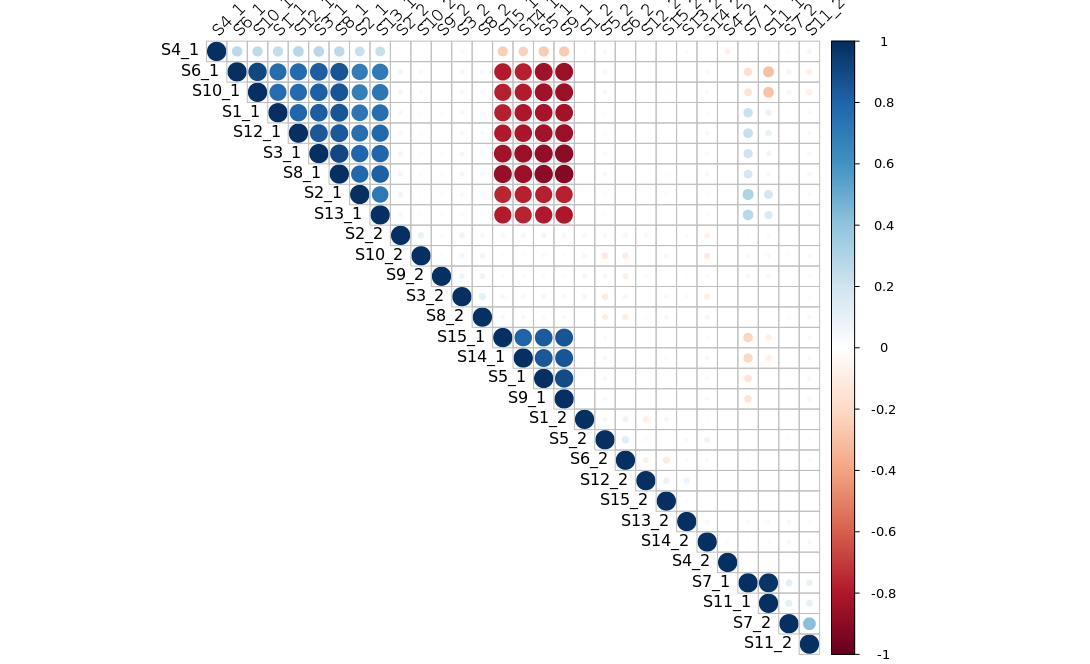}
\vspace{-0.01in}
\caption{Correlation matrix for the projection scores of sensors calculated from the univariate PACE. }\label{corr_socre}
\end{figure*}

\begin{figure*}[htbp]
	\centering
	\includegraphics[width=7.75cm]{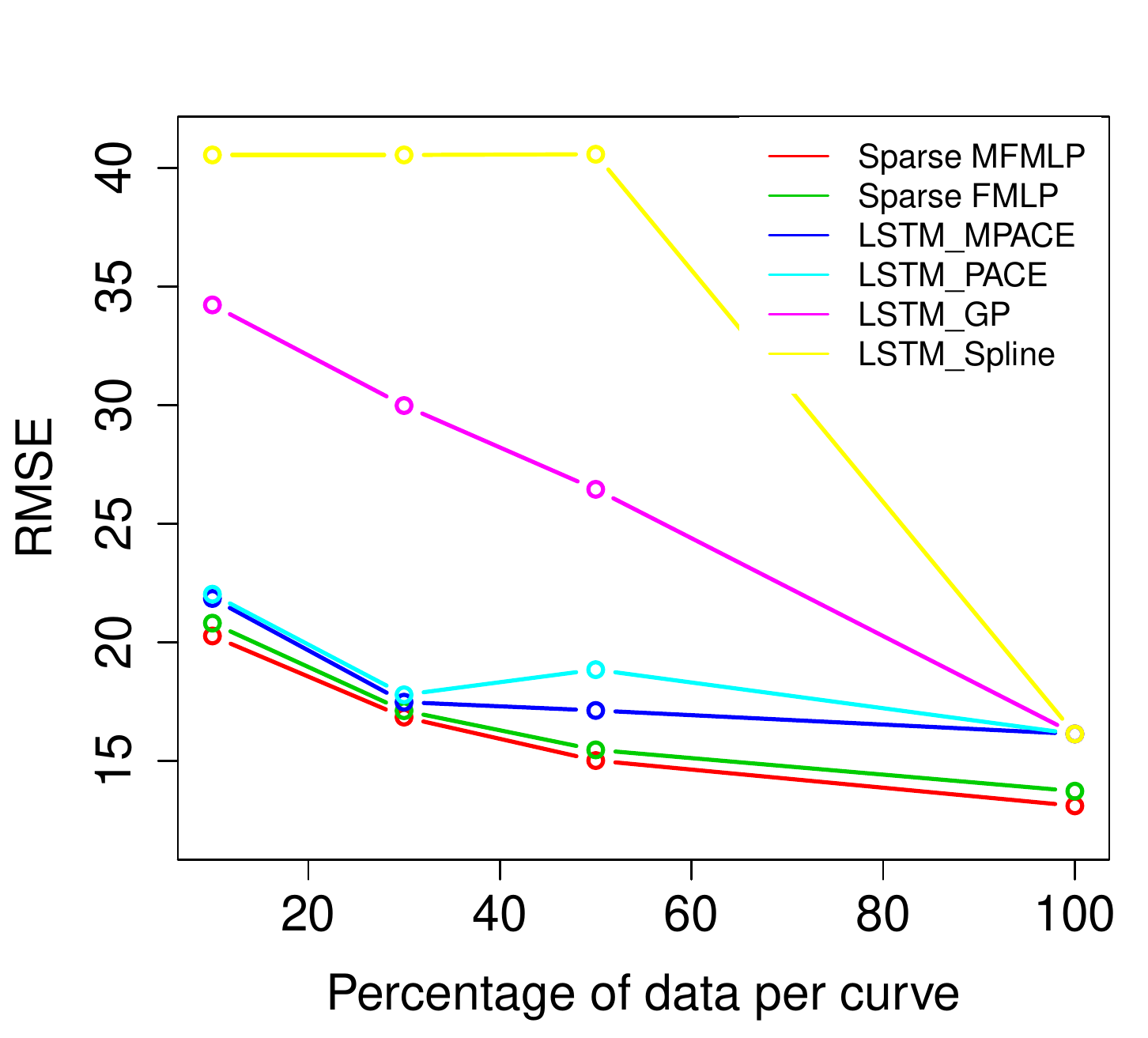}
\vspace{-0.01in}
\caption{RMSE under different percentage of data per subject. }\label{Trend_rmse}
\end{figure*}

For each of the 21 sensor variables, the correlations among the projection scores from separately conducting FPCA on different sensors are visualized in Figure \ref{corr_socre}. It can be seen that there are strong correlations among some of the sensors. Therefore, there will be some benefits of considering the multivariate FPCA in Section \ref{sec2.4.1}. The architecture of the sparse functional MLP (`Sparse FMLP') and the multivariate sparse functional MLP (`Sparse MFMLP') we used is that there are four functional neurons on the first layer, i.e., $K=4$ in \eqref{sparse55} and \eqref{sparse6}, followed by another layer of two numerical neurons. The activation functions in both layers are logistic function. And the weight function is the same as Eq.\eqref{exp1}. The basis functions in Eq.\eqref{exp1} are the eigenfunction from separate FPCA for `Sparse MFMLP' and the eigenfunctions from multivariate FPCA for `Sparse MFMLP'. The baselines respectively utilize the cubic spline, Gaussian process regression, PACE and multivariate PACE to prepare data for LSTM (`$\text{LSTM}_\text{Spline}$', `$\text{LSTM}_\text{GP}$', `$\text{LSTM}_\text{PACE}$', `$\text{LSTM}_\text{MPACE}$'). The RMSE are summarized in Table.~\ref{exp3_tab1}. The plots of RMSE over different level of data sparsity for the considered approaches are provided in Figure \ref{Trend_rmse}. The plot is produced with the result for data set 'FD001'.

Here a summary of the major observations. First, the performance of LSTM varies across different interpolation techniques. What's more, the LSTM with FDA type of interpolation approaches significantly outperforms the conventional data interpolation techniques. This is because the fitting error of the FDA type of interpolations is smaller. Second, the proposed models in the paper outperform all the LSTM based approaches. Specifically, the proposed models is more accurate the `$\text{LSTM}_\text{PACE}$' and `$\text{LSTM}_\text{MPACE}$'. This indicates that the functional way of modeling is better than the sequential calculation in LSTM for this specific problem. Third, the multivariate FMLP performs slightly better than the FMLP based on separate PACE. This is consistent with our intuition, as there are some correlations among the projection scores and `Sparse MFMLP' is proposed to handle such correlations. Note that the magnitude of improvement is not large. We think it is reasonable as the correlation exists among small clusters of projections. As another point, under the dense data scenario, i.e., $100\%$ for both training and testing, our proposed sparse FMLP and the dense FMLP \cite{rossi2002functional} has comparable performance. This experimentally justifies their equivalence under dense data circumstance as discussed in Section \ref{sec2.4.1}. Last, as shown in Figure \ref{Trend_rmse}, the performance of models utilize functional thinking is less affected when the number of data points per curve decreases, compared to models such as `$\text{LSTM}_\text{Spline}$', `$\text{LSTM}_\text{GP}$'.

%The architecture of the sparse functional MLP is that there are four functional neurons on the first layer followed by a layer of two numerical neurons. The accuracy results from our proposed sparse functional MLP and LSTM are given in Table.~\ref{exp3_tab1}. It can be seen that, given a level of sparsity, our proposed sparse functional MLP is performing better than LSTM in terms of both RMSE and score for all the four data sets. Additionally, with the increase of the number of observations per trajectory, both the proposed sparse functional MLP and LSTM get better performances, which is due to the increased amount of information available regarding each engine. Under the dense data scenario, i.e., $100\%$ for both training and testing, our proposed sparse FMLP and the dense FMLP \cite{rossi2002functional} has comparable performance. This experimentally justifies their equivalence discussed in Section \ref{sec2.4}. These two functional MLP outforms the previous best results from LSTM in \cite{zheng2017long}. 

\begin{table}[htbp]
\caption{RMSE comparison on C-MAPSS data.}
\begin{center}
%\scalebox{0.9}{
\begin{tabular}{c|c|c|c c c c}
\hline
\hline
\textbf{Train} & \textbf{Test} & \textbf{Model}& \textbf{FD001}&   \textbf{FD002}&  \textbf{FD003}& \textbf{FD004}\\
\hline
%$30\%$& $30\%$& LSTM& 40.54 & 43.76& 39.81 &   44.28 \\
 %&  & LSTM (GP)&  &  &   &    \\
 %&  & LSTM (PACE)& 26.64 & 31.63 & 25.51  &    \\
 % &  & LSTM (MPACE)& 24.17 & 31.54 & 25.58  &    \\
 % &  & GRU (PACE)&  &  &   &    \\
% &  & Sparse FMLP& \textbf{16.22} & \textbf{19.15} & \textbf{15.40}  & \textbf{20.30}   \\
%\hline
$10\%$& $100\%$& $\text{LSTM}_\text{Spline}$& 40.54 & 43.76 & 39.81 & 49.02   \\
&  & $\text{LSTM}_\text{GP}$ & 34.22  & 36.78 &  35.89 &  34.33  \\
&  & $\text{LSTM}_\text{PACE}$ & 22.03 & 23.69 & 21.89  & 23.75   \\
&  & $\text{LSTM}\text{MPACE}$& 21.85 & 23.10 & 21.10  &  23.54  \\
%  &  & GRU (PACE)&  &  &   &    \\
 &  & Sparse FMLP& 20.81 & 19.53 & 19.83  & 19.39   \\
  &  & Sparse MFMLP& \textbf{20.27} & \textbf{18.67} & \textbf{19.63}  & \textbf{19.03}   \\
\hline
$30\%$& $100\%$& $\text{LSTM}_\text{Spline}$& 40.54 & 43.76 & 39.81 & 49.02   \\
&  & $\text{LSTM}_\text{GP}$& 29.98  & 30.28 & 30.42  &  28.91  \\
&  & $\text{LSTM}_\text{PACE}$& 17.79 & 18.94 & 20.26  &   21.64 \\
&  & $\text{LSTM}_\text{MPACE}$& 17.48 & 18.36 & 20.09  &  21.56  \\
%  &  & GRU (PACE)&  &  &   &    \\
 &  & Sparse FMLP& 17.12 & 17.45 & 16.41  & 18.34   \\
  &  & Sparse MFMLP& \textbf{16.85} & \textbf{17.08} & \textbf{16.06}  & \textbf{18.18}   \\
\hline
%$50\%$& $50\%$  & LSTM& 40.57 & 43.76 & 39.81 &   44.28 \\
%&  & LSTM (GP)&  &  &   &    \\
%&  & LSTM (PACE)& 22.95 &  & 24.24  &    \\
 % &  & GRU (PACE)&  &  &   &    \\
% &&Sparse FMLP& \textbf{14.27} & \textbf{17.22} & \textbf{14.13}  & \textbf{18.35}  \\
%\hline
$50\%$ & $100\%$& $\text{LSTM}_\text{Spline}$& 40.57 & 49.56 & 39.82  & 44.35 \\
&  & $\text{LSTM}_\text{GP}$& 26.45 & 24.76 & 26.43  &   25.26 \\
&  & $\text{LSTM}_\text{PACE}$& 18.85 & 18.84 & 20.44  &   20.77 \\
&  & $\text{LSTM}_\text{MPACE}$& 17.13 & 18.52  & 19.49  &  20.14  \\
%  &  & GRU (PACE)&  &  &   &    \\
 & &Sparse FMLP& 15.47 & 17.24 & 14.63  & 17.44  \\
  &  & Sparse MFMLP& \textbf{15.02} & \textbf{16.98} & \textbf{14.41}  & \textbf{17.19}   \\
\hline
\hline
\end{tabular}

\bigskip

\begin{tabular}{c|c|c|c c c c}
%\caption{RMSE comparison on C-MAPSS data.}
\hline
\hline
\textbf{Train} & \textbf{Test} & \textbf{Model}& \textbf{FD001}&   \textbf{FD002}&  \textbf{FD003}& \textbf{FD004}\\
\hline
%\%$ & $100\%$&MLP\cite{babu2016deep}& 37.56 & 80.03 & 37.39  & 77.37  \\
%&&SVR\cite{babu2016deep} & 20.96 & 42.00 & 21.05 &  45.35 \\
 $100\%$ & $100\%$ &LSTM \cite{zheng2017long} & 16.14 & 24.49& 16.18 &   28.17 \\
%&  & GRU &  &  &   &    \\
& &FMLP& 13.36 & 16.62 & 12.74  & 17.76  \\
 & &Sparse FMLP& 13.73 & 17.04 & 12.75  & 16.92  \\
 & &Sparse MFMLP& \textbf{13.11} & \textbf{16.03} & \textbf{11.97}  & \textbf{16.33}  \\
\hline
\hline
\end{tabular}
%}
\label{exp3_tab1}
\end{center}
\end{table}

\section{Conclusion and Discussion}
\label{sec5}
In this paper, we focused on the temporal classification/regression problem, the purpose of which is to learn a mathematical mapping from the time series inputs to scalar response, leveraging the temporal dependencies and patterns. In real-world applications, we noticed that two types of data scarcity are frequently encountered: scarcity in terms of small sample sizes and scarcity introduced by sparsely and irregularly observed time series covariates. Noticing the lack of feasible temporal predictive models for sparse time series data in the literature, we proposed two sparse functional MLP (`Sparse FMLP' and `Sparse MFMLP') to specifically handle this problem. The proposed SFMLP is an extension of the conventional FMLP for densely observed time series data, employing the univariate and multivariate sparse functional principal component analysis. We used mathematically arguments and numerical experiments to evaluate the performance of each candidate model under different types of data scarcity and achieved the following conclusions:

\begin{itemize}[leftmargin=*]
\item When the sample size is large and the underlying function that gives rise to the time series observations is non-smooth over time, the sequential learning models are more appropriate models to handle the supervised learning task.  
\item When the sample size is small and the underlying function is smooth, we expect FMLP or equivalently the proposed SFMLP to outperform the sequential learning in general, since
\begin{enumerate}
    \item FMLP requires fewer training samples due to its feed-forward network structure, succinct temporal pattern capturing technique, and its capability of encoding domain expert knowledge.
    \item FMLP is easier to train as the time-varying impact of the covariates on the response is aggregated through integrals of the continuous processes, rather than recursive computations on the individual observations. 
    \item FMLP is less restrictive on the data format. In particular, the time series covariates can be regular or irregular. Also, the number of observations as well as the measuring timestamps can be different across features and subjects.
\end{enumerate}
\item The proposed sparse FMLPs are feasible solutions when the individual time series features are sparsely and irregularly evaluated.
\end{itemize}

\bibliography{fanova}

\end{document}